\documentclass[10pt,journal,compsoc]{IEEEtran}
\usepackage{graphicx}
\usepackage{xcolor}

\usepackage{color,soul}

\soulregister\cite7
\soulregister\ref7
\soulregister\pageref7

\ifCLASSOPTIONcompsoc
  \usepackage[nocompress]{cite}
\else
  \usepackage{cite}
\fi
\ifCLASSINFOpdf
\else
\fi

\definecolor{chocolate}{RGB}{210,105,30}

\begin{document}
%
\title{Multi-Scale Networks for 3D Human Pose Estimation with Inference Stage Optimization}

%
%
%
%

\author{Cheng~Yu,
        Bo~Wang,
        Bo~Yang,
        and~Robby~T.~Tan,~\IEEEmembership{Member,~IEEE}
        
\IEEEcompsocitemizethanks{\IEEEcompsocthanksitem ChengYu and Robby T. Tan are with Yale-NUS College and the Department of Electrical
and Computer Engineering, National University of Singapore, Singapore,
(e-mail: e0321276@u.nus.edu, robby.tan@nus.edu.sg).

\IEEEcompsocthanksitem Bo Wang and Bo Yang are with Tencent America, Santa Monica, CA, USA, (e-mail: \{bohawkwang, brandonyang\}@tencent.com).}

\thanks{Manuscript received July 25, 2020.}}

%
%

\markboth{Journal of \LaTeX\ Class Files,~Vol.~14, No.~8, August~2015}%
{Shell \MakeLowercase{\textit{et al.}}: Bare Demo of IEEEtran.cls for Computer Society Journals}
%



\IEEEtitleabstractindextext{%
\begin{abstract}
Estimating 3D human poses from a monocular video is still a challenging task. Many existing methods' performance drops when the target person is occluded by other objects, or the motion is too fast/slow relative to the scale and speed of the training data. Moreover, many of these methods are not designed or trained under severe occlusion explicitly, making their performance on handling occlusion compromised. Addressing these problems, we introduce a spatio-temporal network for robust 3D human pose estimation. As humans in videos may appear in different scales and have various motion speeds, we apply multi-scale spatial features for 2D joints or keypoints prediction in each individual frame, and multi-stride temporal convolutional networks (TCNs) to estimate 3D joints or keypoints. Furthermore, we design a spatio-temporal discriminator based on body structures as well as limb motions to assess  whether the predicted pose forms a valid pose and a valid movement. During training, we explicitly mask out some keypoints to simulate various occlusion cases, from minor to severe occlusion, so that our network can learn better and becomes robust to various degrees of occlusion. As there are limited 3D ground-truth data, we further utilize 2D video data to inject a semi-supervised learning capability to our network. Moreover, we observe that there is a discrepancy between 3D pose prediction and 2D pose estimation due to different pose variations between video and image training datasets. We therefore propose a confidence-based  inference stage optimization to adaptively enforce 3D pose projection to match 2D pose estimation to further improve final pose prediction accuracy. Experiments on public datasets validate the effectiveness of our method, and our ablation studies show the strengths of our network's individual submodules.
\end{abstract}

\begin{IEEEkeywords}
3D human pose estimation, occlusion, convolutional neural network, temporal convolutional network.
\end{IEEEkeywords}}

\maketitle

\IEEEdisplaynontitleabstractindextext

%
\IEEEpeerreviewmaketitle

\IEEEraisesectionheading{\section{Introduction}\label{sec:introduction}}

\label{sec:intro}
This paper focuses on 3D human pose estimation from a monocular RGB video. A 3D pose is defined as the 3D coordinates of pre-defined keypoints on humans, such as shoulder, pelvis, wrist, and etc. Recent top-down approaches \cite{hossain2018exploiting,Repnet,pavllo20183d,ChengICCV19} have shown promising results, where spatial features from individual frames are extracted to detect a target person and estimate the 2D poses, and temporal context is used to produce consistent 3D human pose predictions. 
However, we find that existing methods do not fully exploit the spatial and temporal information in videos. Single-image based methods focus on using spatial information but ignores the temporal information; video based methods try to employ temporal information in single scale and only use single-frame based discriminator to check pose validity which cannot guarantee the estimated pose sequence is valid.
As a result, they suffer from the problem of large variations in sizes and speeds of the target person in wild videos.

\begin{figure}[t]
\centering
\includegraphics[width=\linewidth]{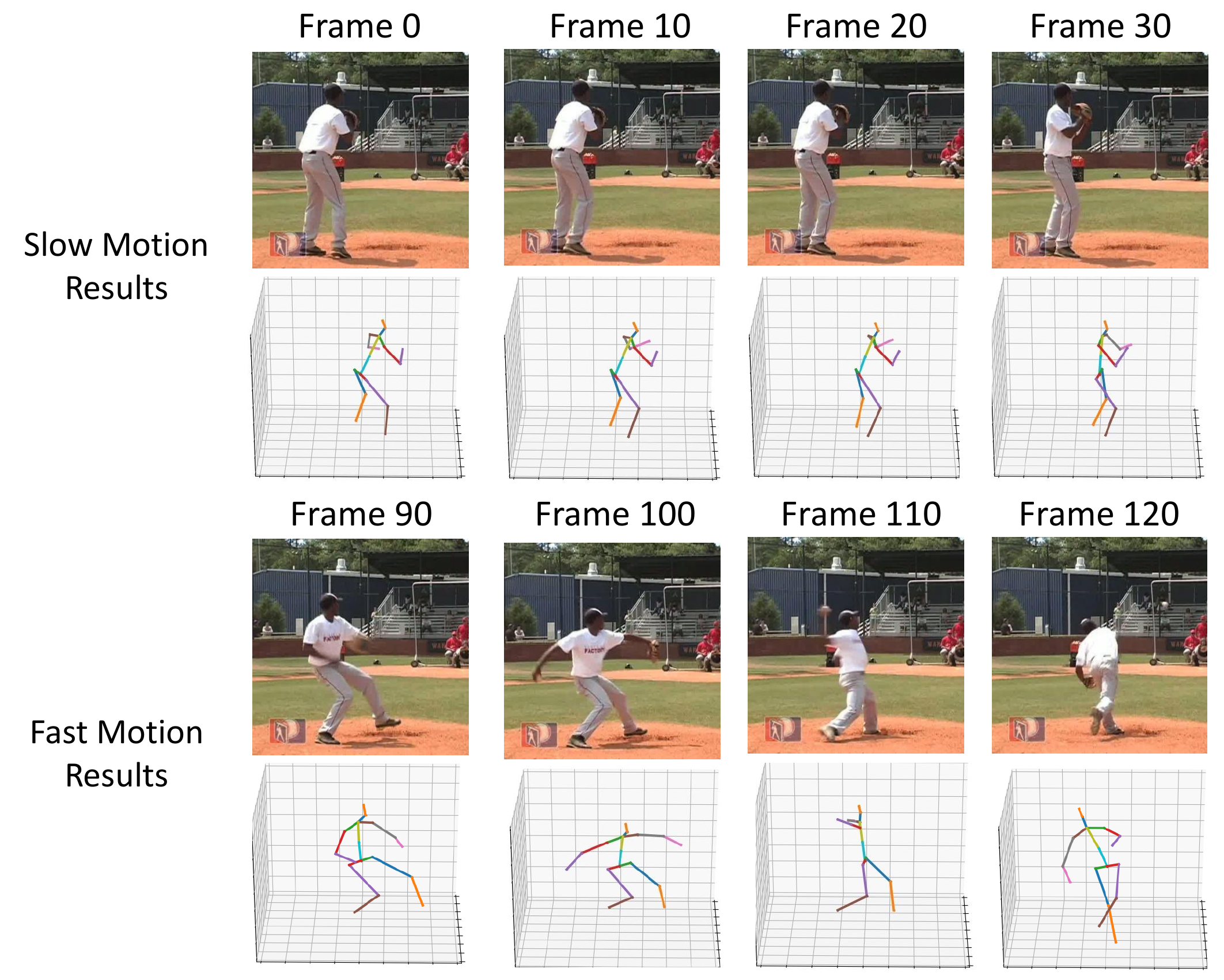}
\caption{Examples of our 3D human pose estimation under different movement speeds.} 
\label{fig:multi-scale-example}
\end{figure}

Addressing the problem, we consider multi-scale features both spatially and temporally to deal with persons at various distances with different speeds of motions. We use the High Resolution Network (HRNet) \cite{sun2019deep}, which makes use of multi-scale spatial features to produce one heatmap for each keypoint. Unlike most previous methods (e.g.~\cite{newell2016stacked,pavllo20183d}) that use only the peaks in the heatmaps, we encode these maps into a latent space to incorporate more spatial information. We then apply temporal convolutional networks (TCNs)~\cite{pavllo20183d} to these latent features with different strides,  and concatenate them together for prediction of the 3D poses. Figure~\ref{fig:multi-scale-example} shows some examples of our results. 

\begin{figure*}[t]
\centering
\includegraphics[width=\linewidth]{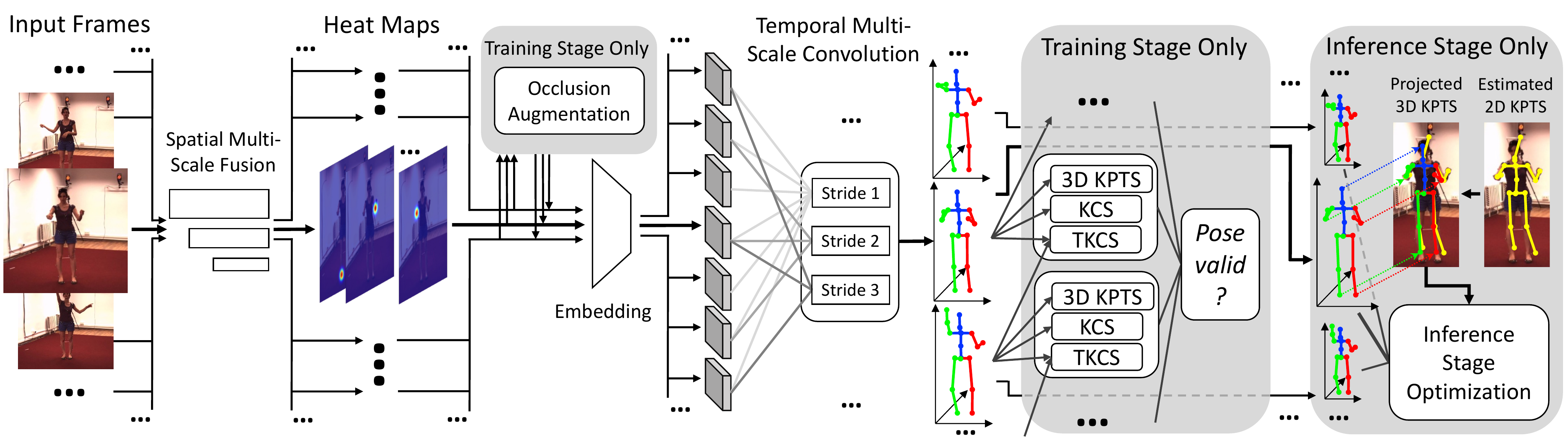}
\caption{Illustration for our framework. We only show three different temporal strides for clarity purpose. KPTS is short for keypoints; KCS is Kinematic Chain Space; TKCS means Temporal KCS.} 
\label{fig:framework}
\end{figure*}

Moreover, to reduce the risk of predicting invalid 3D poses, we utilize a discriminator in our framework like previous methods \cite{Yang20183DHP,Repnet,Chen_2019_CVPR,zhang2019phd}. Unlike these methods, we check the pose validity spatio-temporally. The main reason is that valid poses in individual frames do not necessarily constitute a valid sequence. We extend the spatial KCS (Kinematic Chain Space)~\cite{Repnet},  and introduce a temporal KCS to represent motions of human joints. This temporal KCS descriptor is used by another TCN to check the validity of the estimated 3D pose sequence.

Furthermore, to deal with occlusion, during the training of our TCNs, we mask out some keypoints or frames by setting the corresponding heatmaps to zero, as shown in Figure \ref{fig:framework}. There are two types of methods that are similar to ours. One is the partial occlusion modeling by setting coordinates of some keypoints to zero~\cite{ChengICCV19}. The other is human dynamics, which only handles occlusion that happens in the end of a temporal window, since it predicts several future frames from given past frames' information~\cite{humanMotionKanazawa19,zhang2019phd}.
Unlike these methods, our method can handle both partial and total occlusion cases in individual frames or in a sequence of frames. Thus, it is more general in handling human 3D pose estimation under occlusion. Besides, our occlusion module also allows us to do semi-supervised learning utilizing both 3D and 2D video datasets.

As our TCNs are trained on video data and their 3D ground truths, which have limited pose variations compared with 2D image datasets~\cite{li2020cascaded}, the output 3D poses of our TCNs can be erroneous, despite our correct estimations of the 2D keypoints.
To address this, we introduce a novel confidence-based optimization to better enforce the consistency between 3D pose prediction and 2D pose estimation in the inference stage, which we call inference stage optimization (ISO).
The proposed ISO is aware of the error in 2D pose estimations and adaptively adjusts how much the projected 3D pose should match the 2D pose estimation based on the confidence of 2D pose estimator. It is able to leverage the correctly predicted keypoints of 2D pose estimation in a self-supervised way, therefore, boosting the 3D pose estimation accuracy in the test time.

This paper is a substantial extension of our previous conference papers~\cite{cheng2020sptaiotemporal} and cylinder man model~\cite{ChengICCV19} with additional parameter analysis, comparisons, and new inference stage optimization (ISO) that further improves the 3D pose estimation accuracy. As a summary, our contributions are as follows:
\begin{itemize}
    \item Incorporate multi-scale spatial and temporal features for robust pose estimation in video.
    \item Introduce a spatio-temporal discriminator to regularize the validity of a pose sequence.
    \item Perform diverse data augmentation for TCN to deal with different occlusion cases.
    \item Propose a self-supervised inference stage optimization to enhance the 3D pose estimation accuracy.
\end{itemize}
Experiments on public datasets show the efficacy of our method.

\section{Related Works}

Within the last few years, human pose estimation has been undergoing rapid development with deep learning techniques~\cite{tompson2014joint,toshev2014deeppose,pfister2015flowing,carreira2016human,bulat2016human,newell2016stacked,wei2016cpm,zhou2016sparseness,chu2017multi,cao2017realtime,mehta2017vnect,martinez2017simple,belagiannis2017recurrent,chen20173d,yang2017learning,mehta2017monocular,chen2017adversarial,cao2018openpose,alp2018densepose,xiao2018simple,Yang20183DHP,pavlakos2018ordinal,sun2018integral,ke2018multi,Li_2019_CVPR,Chen_2019_CVPR}. Researchers keep pushing the frontier of this field from different angles via better utilizing spatial or temporal information, learning human dynamics, pose regularization, and semi-supervised/self-supervised learning~\cite{sun2019deep,hossain2018exploiting,humanMotionKanazawa19,Repnet,pavllo20183d}. 

\vspace{2mm}
\noindent
\textbf{Utilizing spatial and temporal information.}
To better utilize spatial information, some recent methods focused on cross stage feature aggregation or multi-scale spatial feature fusion to maintain the high resolution in the feature maps~\cite{chen2018cascaded,sun2019deep,humanMotionKanazawa19}. Although this helps to improve the 2D estimators, there is an inherent ambiguity for inferring 3D human structure from a single 2D image. To overcome this limitation, some researchers further utilized temporal information in video~\cite{pavllo20183d,hossain2018exploiting,ChengICCV19,bertasius2019learning}, and showed obvious improvement. However, their fixed temporal scales limit their performance on videos with different motion speeds from the ones in training. 

\vspace{2mm}
\noindent
\textbf{Regularizing estimated human pose.}
To regularize predictions to be reasonable 3D human poses, pose discriminators have been proposed~\cite{Yang20183DHP,Repnet,Chen_2019_CVPR}. These methods utilize the idea of Generative Adversarial Networks (GAN)~\cite{goodfellow2014gan} to check whether the estimated 3D pose is consistent with the pose distribution in the ground-truth data. However, most of existing methods only focus on determining if one given 3D human pose is reasonable. Combining a series of reasonable 3D poses together does not make the whole series a reasonable human motion trajectory. As a result, we propose temporal KCS which checks both the spatial and temporal validity of 3D poses. 

\vspace{2mm}
\noindent
\textbf{Handling different types of occlusion.}
To deal with partial occlusions, some techniques have been designed to recover occluded keypoints from unoccluded ones according to the spatial or temporal context \cite{radwan2013monocular,rogez2017lcr,de2018deep,guo2018occluded,ChengICCV19} or scene constraints~\cite{zanfir2018monocular,zanfir2018deep}. Some methods further introduced the concept of ``human dynamics''~\cite{humanMotionKanazawa19,zhang2019phd}, which predicts future human poses based on single or multiple existing frames in a video without any future frames. In real scenarios, we may have full, partial, or total occlusion for individual or continuous frames. Therefore, we introduce an approach to integrate these two categories of methods into one unified framework by explicitly performing augmentation for different occlusion cases during training.
Due to limited 3D human pose data, recent methods suggest to further utilize 2D human pose datasets in a semi-supervised or self-supervised fashion \cite{Repnet,wang2019selfsupervised,Kocabas_2019_CVPR,Chen_2019_CVPR}. They project estimated 3D pose back to 2D image space so that 2D ground-truth can be used for loss computation. Such approaches reduce the risk of over-fitting on small amount of 3D data. We also adopt this method and combine it with our explicit occlusion augmentation. 

\vspace{2mm}
\noindent
\textbf{Inference stage optimization.}
Recently, self-supervised inference stage (test time) optimization has been proposed and applied to different computer vision or machine learning applications such as point cloud reconstruction~\cite{navaneet2020image}, image super resolution~\cite{chen2019single}, and motion capture~\cite{tung2017self}. Among these method, they rely on directly using input image to preserve low-resolution features~\cite{chen2019single}, differentiable rendering~\cite{tung2017self}, or image projection difference~\cite{navaneet2020image} to achieve test time optimization. Unfortunately, none of these approaches are applicable to 3D pose estimation as we are dealing with estimation of a set of keypoints. We propose to leverage the 2D pose estimation information in a self-supervised way to achieve test time optimization for 3D human pose estimation. 
Due to the 2D pose estimation is not always correct, our inference stage optimization is designed to be dependent on the confidence scores of the 2D estimator, making our inference stage optimization more robust to 2D pose estimation errors.

\section{Proposed Method}

Given an input video, we first detect and track the persons by any state-of-the-art detector and tracker, such as Mask R-CNN \cite{he2017maskrcnn} and PoseFlow \cite{xiu2018poseflow}. Subsequently, we perform the pose estimation for each person individually.

\subsection{Multi-Scale Features for Pose Estimation}
Given a series of bounding boxes for a person in a video, we first normalize the image within each bounding box to a pre-defined fixed size, e.g., $256 \times 256$, and then apply High Resolution Networks (HRNet) \cite{sun2019deep} to each normalized image patch to produce $K$ heatmaps, each indicates the possibility of certain human joint's location. The HRNet conducts repeated multi-scale
fusions by exchanging the information across the parallel multi-scale subnetworks. Thus, the estimated heatmaps incorporate spatial multi-scale features to provide more accurate 2D pose estimations. We follow the training pipeline of 2D pose estimator \cite{de2018deep}, where ground truth for visible joints are rendered as Gaussian heatmaps and ground truth for invisible joints are zero heatmaps. As the 3D pose datasets are obtained from motion capture sequences, which does not contain visibility labels, we obtain pseudo-occlusion label using Cylinder Man Model which is described as in Section~\ref{sec:cylindman}. MPII~\cite{andriluka20142d} is used for HRNet training, and Human3.6M~\cite{h36m_pami} is used as well for Human3.6M related evaluations.

We concatenate the $K$ heatmaps in each frame as a $K$-dimensional image $m_t$, where $t$ is the frame index, and apply an embedding network $f_E$ to produce a low dimensional representation as $r_t=f_E(m_t)$. Such embedding incorporates more spatial information from the whole heatmaps than only using maps' peaks as most previous methods do. The effectiveness of the embedding is shown in the ablation study in Table~\ref{tab:ablation}. 

\begin{figure}
    \centering
    \includegraphics[width=\linewidth]{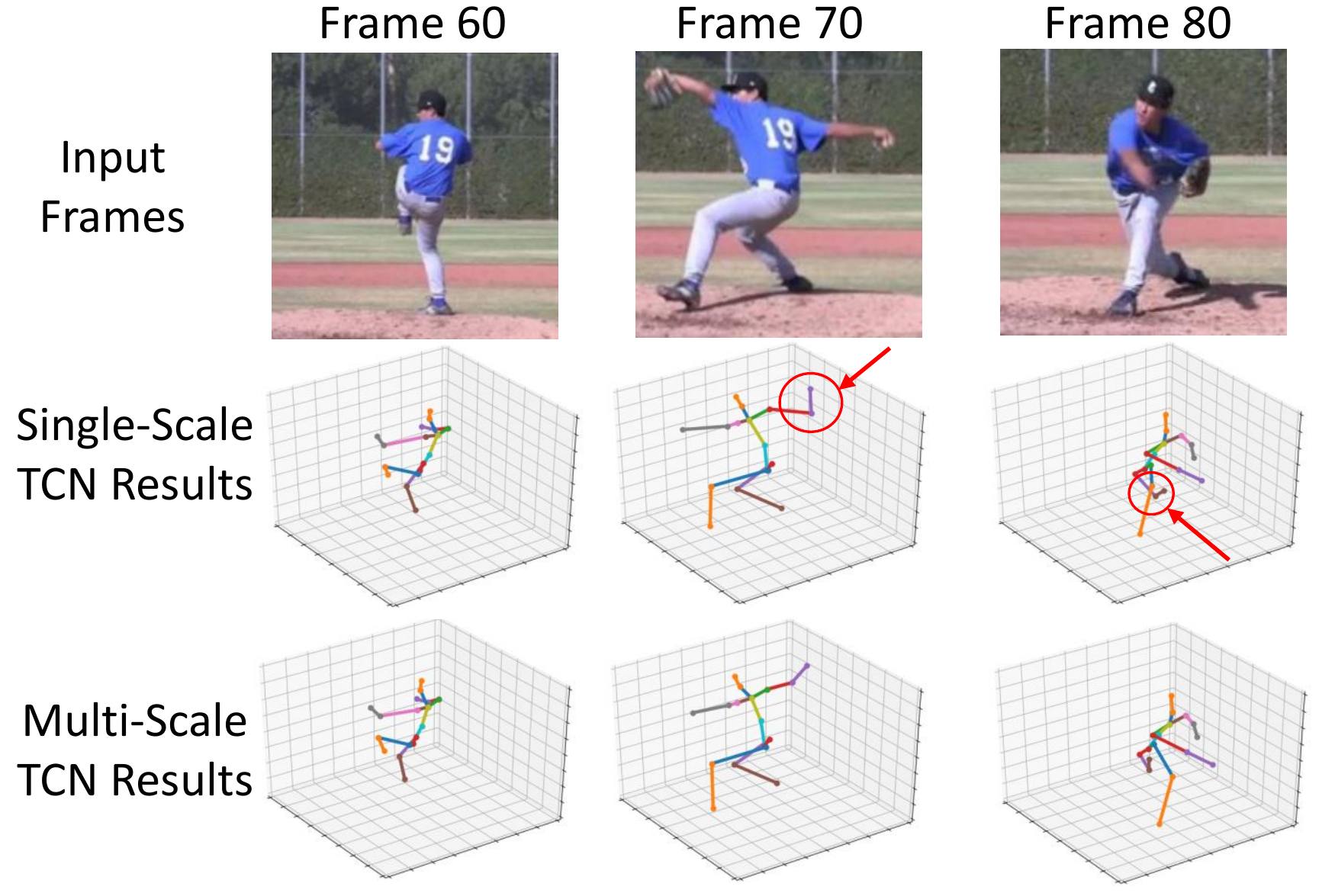}
    \caption{Comparison of single-scale and multi-scale TCN results. Errors are labeled in red circles. The single-scale TCN fails to provide accurate predictions for fast motion frames.} 
    \label{fig:multiscale}
\end{figure}

Given a sequence of heatmap embeddings $\{r_t\}$, we apply TCN to them. As human motions may be fast or slow, we consider multi-scale features in the temporal domain. As shown in Figure \ref{fig:framework}, we apply TCN with temporal strides of 1, 2, 3, 5, 7 and concatenate these features for the final pose estimation. Such multi-scale features in both spatial and temporal domains enable our networks to deal with various scenarios. Figure~\ref{fig:multiscale} shows an example video clip with fast motion of playing baseball. We observe that multi-scale TCN is able to produce more accurate results than single-scale TCN.

We use both 3D dataset Human3.6M~\cite{h36m_pami} and 2D dataset Penn Action~\cite{penndataset} for training. Human3.6M has multi-view captured videos and 3D ground-truth, while PENN only has 2D ground-truth for visible keypoints. For Human3.6M data, the 3D MSE loss is defined as:
\begin{equation}
    L_{3d} = (\tilde{X} - X^{3D})^2,
\end{equation}
where $\tilde{X}$ is our predicted 3D coordinates for all keypoints, and $X^{3D}$ is the 3D ground truth. As Human3.6M data set provides videos from multiple views, we expect the 3D estimation results from different views should be the same after rotation alignment. So, we define the multi-view loss as:

\begin{equation}
    L_{mv} = (R_{v1 \rightarrow v2} \tilde{X}_{v1} - \tilde{X}_{v2} )^2,
\end{equation}
where $R_{v1 \rightarrow v2}$ is the rotation matrix from viewpoint 1 to viewpoint 2, and is precomputed from the ground-truth camera parameters. The $\tilde{X}_{v1}$ and $\tilde{X}_{v2}$ are the predicted 3D results in viewpoints 1 and 2.

For the 2D dataset, we project the 3D prediction to 2D space assuming orthogonal projection, and the 2D MSE loss is defined as: 
\begin{equation}
    L_{2d} = (Orth(\tilde{X}) - X^{2D})^2,
\label{eq:l2d}
\end{equation}
where $Orth(\cdot)$ is the orthogonal projection operator, and $X^{2D}$ is the 2D ground truth.

\subsection{Spatio-Temporal KCS Pose Discriminator}
\label{sec:discriminator}
To reduce the risk of generating unreasonable 3D poses, we introduce a novel spatio-temporal discriminator to check the validity of a pose sequence, rather than just poses in individual frames like previous methods~\cite{Yang20183DHP,Repnet,Chen_2019_CVPR}.

Kinematic Chain Space (KCS) is an effective single frame pose discriminator as demonstrated in~\cite{Repnet}. Each bone, defined as the connection between two neighboring human keypoints such as elbow and wrist, is represented as a 3D vector $b_m$, indicating the direction from one keypoint to its neighbor. All such vectors form a $3 \times M$ matrix $B$, where $M$ is the predefined number of bones for a human structure. $\Psi = B^T B$ is used as the features for discriminator in KCS, where the diagonal elements in $\Psi$ indicate the square of bone length and other elements represent the weighted angle between two bones as an inner production. Please note that KCS only takes spatial configuration into consideration, therefore, it is not capable of checking if a pose is valid temporally. 

\begin{figure}[t]
\centering
\includegraphics[width= \linewidth]{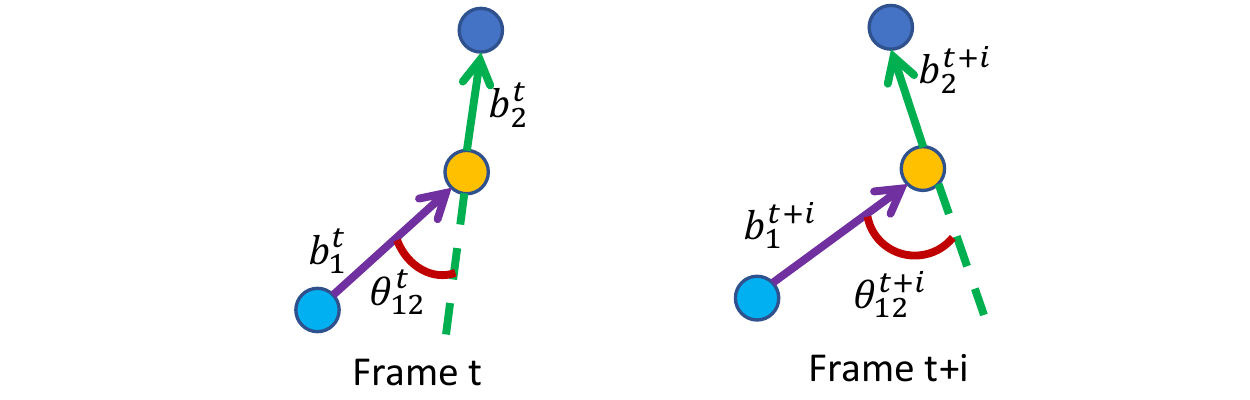}
\caption{Illustration for Temporal Kinematic Chain Space (TKCS) between two neighboring bones.} 
\label{fig:TKCS}
\end{figure}

To better leverage temporal information and overcome the intrinsic weakness of single frame based pose discriminators like KCS, we introduce a Temporal KCS (TKCS) defined as, 
\begin{equation}
    \Phi = B_{t+i}^TB_{t+i} - B_t^TB_t.
\end{equation}
where $i$ is the temporal interval between the KCS. The diagonal elements in $\Phi$ indicates the bone length changes, and other elements denote the change of angles between two bones. 
Figure \ref{fig:TKCS} shows an example of two neighboring bones $b_1$ and $b_2$. The spatial KCS measures the lengths of $b_1$ and $b_2$ as well as angles between them, $\theta_{12}$. The temporal KCS measures the bone length changes between two frames with temporal interval $i$, i.e., differences between $b_1^t$ and $b_1^{t+i}$ as well as $b_2^t$ and $b_2^{t+i}$, and the angle change between neighboring bones, i.e., difference between $\theta_{12}^t$ and $\theta_{12}^{t+i}$.

We concatenate the spatial KCS, temporal KCS, and the predicted keypoint coordinates, and then feed them to a TCN to build a discriminator. Such approach not only considers whether a pose is valid in individual frames, but also checks the validity of transitions across frames. We follow the procedure in the standard GAN to train the discriminator~\cite{goodfellow2014gan}, and use it to produce a regularization loss for our predicted poses as $L_{gen}$.

In addition, to increase the robustness under different view angles, we introduce a rotational matrix as an augmentation to the generated 3D pose, as shown in the following equation:
\begin{equation}
    L'_{gen} = L_{gen}(R\tilde{X}),
\end{equation}
where $R$ is a rotational matrix $Rotation(\alpha, \beta, \gamma)$, and $\alpha$, $\beta$, $\gamma$ are rotational angles along $x$, $y$, and $z$ axis, respectively. As the rotational angles along $x$ and $z$ angles should be smaller compared with rotations along $y$ for normal human poses, in our experiments, $\beta$ is randomly sampled from $[-\pi, \pi]$ while $\alpha$ and $\gamma$ are sampled from $[-0.2\pi, 0.2\pi]$. 

The overall loss function for our training is defined as
\begin{equation} 
    L = L_{3d} + w_1 L_{mv} + w_2 L_{2d} + w_3 L'_{gen},
\end{equation}
where $w_1, w_2, w_3$ are set to $0.5$, $0.1$, $0.01$, respectively, and are fixed in all our experiments.

\subsection{Data Augmentation for Occlusions}
To make our approach capable of dealing with different occlusion cases, we perform data augmentation during the training process. We use random masking of keypoints to simulate various occluded conditions. Four types of occlusion are applied in the training process:

\begin{itemize}
\item
\textbf{Discrete point-wise occlusion}: The keypoints in each frame are randomly masked to zero at a probability $p_1$. The purpose is to introduce noise that not all keypoints are constantly giving high response across all video frames.

\item
\textbf{Discrete frame-wise occlusion}: The frames are randomly chosen to be masked to zero at a probability $p_2$. It is introduced to suppress the effect caused by unreliable frames such as motion blur, abnormal exposures, etc.

\item
\textbf{Continuous point-wise occlusion}: Certain keypoints are randomly chosen to be masked at a probability $p_3$ with random length between $2$ and $l$ at random position. This will lead to a result where several keypoints are constantly invisible in the input sequence. The model will learn to recover the missing keypoints with human dynamics. 

\item
\textbf{Continuous frame-wise occlusion}: A mask with random length between $2$ and $l$ is applied to the input sequence at a random position where all the masked frames' heatmaps are set to 0. The model learns to recover the long-term occlusion with human dynamics.
\end{itemize}

In our experiments, we set the $p_1, p_2, p_3, l$ to $0.2, 0.2, 0.2, 40$ respectively. In addition, as the output of 2D pose estimator may be inaccurate or even flipped (i.e., left and right keypoints are swapped), we introduce random noise to the heatmaps of the input sequence. To further improve the robustness under wrong detection cases, the points are randomly shifted or randomly swapped symmetrically. For example, the left knee is swapped with the right knee and the elbow point is sifted by 10 pixels. We expect the trained multi-scale TCN is able to recover the correct 3D pose using context information from partially wrong 2D estimations.

\begin{figure}[t]
\centering
\includegraphics[width=\linewidth]{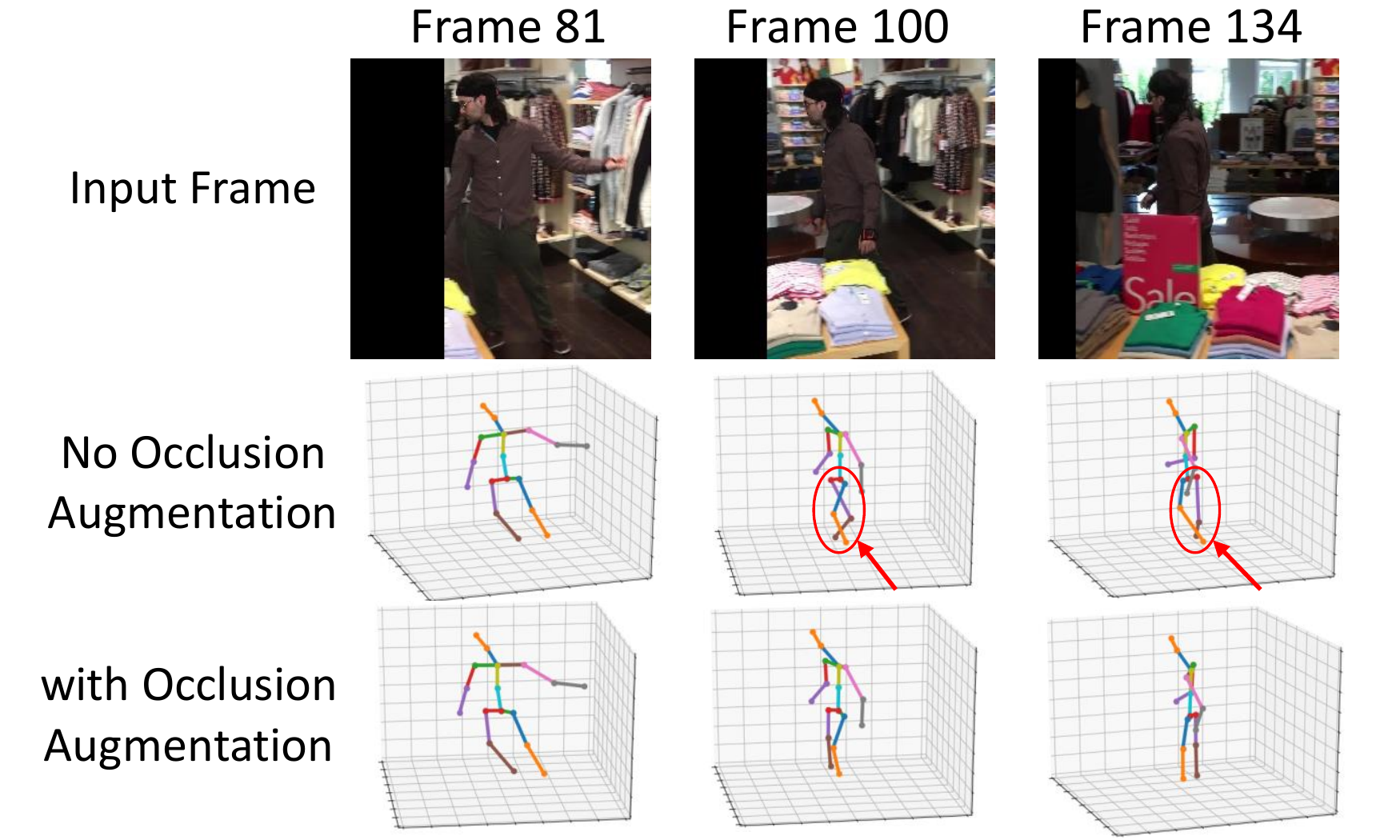}
\caption{Comparison of results from models trained with and without occlusion augmentation. Wrong estimations are labeled in red circles.}
\label{fig:occlusion}
\end{figure}

Note that, when the occlusion masks are all at the end of our TCN receptive field, it degrades to the human dynamics case, i.e., estimation of future poses without any future observation. Our method is able to predict human poses from temporal context information with various kinds of missing or inaccurate observations in a video clip. Therefore, our framework is a more generalized approach for occlusion handling. Figure \ref{fig:occlusion} demonstrates an example where occlusion augmentation helps to generate robust pose estimation results in a video clip where a target person is occluded. 

\subsection{Cylinder Man Model}
\label{sec:cylindman}

Training the 3D TCN requires pairs of a 3D joint ground-truth and a 2D keypoint with occlusion label. However, the existing 3D human pose datasets (e.g., \cite{h36m_pami,mono-3dhp2017}) have no occlusion labels, and the amount of the 3D data is limited. Hence, we introduce a ``Cylinder Man Model'' to generate occlusion labels for 3D data and perform data augmentation.

\begin{figure}[h]
\centering
    \includegraphics[width=0.85\linewidth]{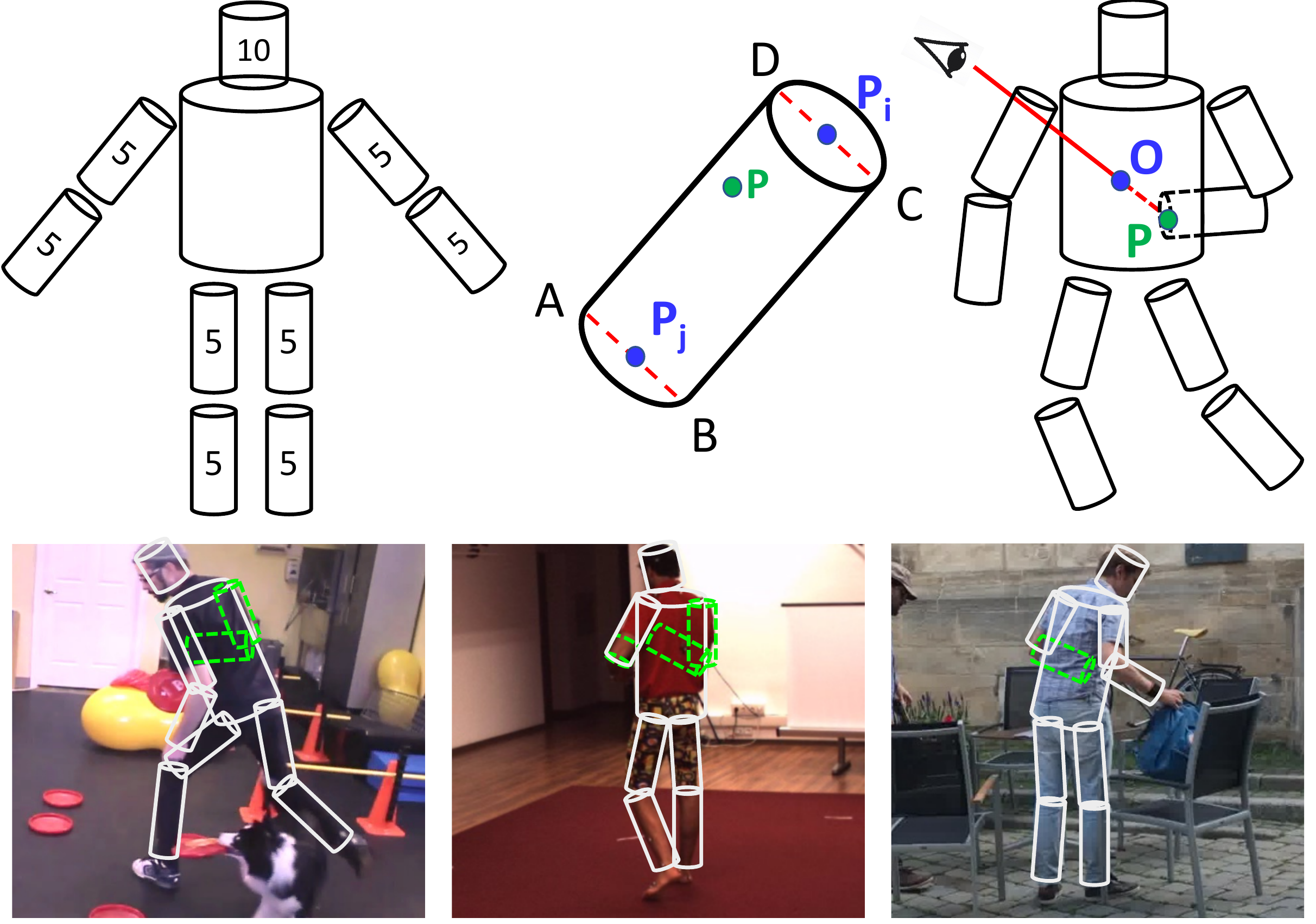}
    \caption{Illustration for the "Cylinder Man Model" for occlusion reasoning. See text for details.}
    \label{fig:cubeman}
\vspace{-1em}
\end{figure}

As shown in the top left of Figure~\ref{fig:cubeman}, we divide a 3D human into ten parts: head, torso, two upper arms, two lower arms, two thighs, and two lower legs. Given any 3D skeleton, either from the ground-truth or our networks, we use a cylinder to approximate the 3D shape of each of the ten parts. The radius of the head is defined as $10cm$, and the radius of each limb is defined as $5cm$ as labeled in Figure~\ref{fig:cubeman} top left. The height of a cylinder is defined as the distance between keypoints defining that part. The radius of the torso is not pre-defined but is set to the distance between the neck and the shoulder. Such approximation works well in our framework, and is validated by the experiments.


In our model, each cylinder is formulated by $C_{ij}=\{r_{ij}, P_i, P_j\}$, where $r_{ij}$ is the radius and $P_i$, $P_j$ are the 3D joints defining the centers of the top and bottom parts of the cylinder as visualized in the top middle of Figure~\ref{fig:cubeman}. 

To calculate whether a point $P$ is occluded by $C_{ij}$, we first map them to the 2D space assuming orthogonal projection. The vertical cross section of the cylinder, $ABCD$, maps to a rectangle $A'B'C'D'$ as shown in the top middle of Figure~\ref{fig:cubeman}. As $r_{ij}$ is small with respect to the height of the cylinder, i.e., the length of a bone, we only check whether $P$ is occluded by $ABCD$ when projecting to the 2D plane. If the projected $P$ is not inside the rectangle $A'B'C'D'$ in 2D space, it is not occluded. Otherwise, we calculate the norm of the plane $ABCD$ in 3D space as $\overrightarrow{n_{ij}}=\overrightarrow{P_j P_j} \times \overrightarrow{P_j A}$. Note that,  $-\overrightarrow{n_{ij}}$ is also the norm vector. We choose the one pointing toward the camera, i.e., $z$ coordinate is negative.

The visibility of point $P$ is then calculated by
\begin{equation}
\label{equ:visibility}
    V_P = \prod_{(i,j) \in E} [(P - P_i) \cdot n_{ij} > 0],
\end{equation}
where $E$ is the set of all neighboring keypoints forming a bone, and $[ \cdot ]$ denotes the Iverson bracket, returning $1$ if the proposition is true, and $0$ otherwise. In order to assure differentiable, we use sigmoid function to approximate this operation. As shown in the top right of Figure \ref{fig:cubeman}, if the view angle is from behind the person, the keypoint $P$ will be occluded by the body cylinder at point $O$. In the bottom of Figure \ref{fig:cubeman}, cylinder man models are overlaid on top of images as an illustration. 

Other human body model like SMPL \cite{SMPL2015} could provide more detailed representation of human shape, but it requires extra computational cost for checking occlusion. Cylinder based approximation is suitable and sufficient for our tasks.

\subsection{Inference Stage Optimization}

Using the common deep learning framework, we train our network on a training dataset to  produce a model. This model is then utilized to predict 3D poses from a testing dataset. 
Doing in this way, however, we found the performance of the model can be erroneous. The main  reason is because there are domain gaps between the training and testing data, particularly for our 3D pose training data, which is more limited compared to our 2D pose training data.
For this reason, we propose to perform a self-supervised optimization in the testing/inference stage to further enhance our method's performance.
We apply a similar re-projection error loss as Eq.~(\ref{eq:l2d}), but since the ground-truth of 2D keypoints is not available in testing videos, the loss term is computed between the orthogonal projection of estimated 3D pose and the output of 2D pose estimator, formulated as:
\begin{equation}
    L_{rep} = \sum_{k=1}^K w_{k} L_{2d}^2 = \sum_{k=1}^K w_{k}(Orth(X_k) - X^{det}_k)^2
    \label{eq:rep}
\end{equation}
where $K$ is the number of keypoints and $w_{k}$ is the weight of each keypoint for re-projection loss. Different ways of setting $w_{k}$ are discussed in the following paragraph.

First, a fixed constant value like $1.0$ can be used as the weight $w_{k}$. The re-projection loss with a constant weight means the 3D pose projection is always enforced to match the 2D pose estimation, regardless whether the 2D estimator is correct. 
Second, the confidence score of 2D pose estimator $c_k$ can be used as the weight as it might reflects the accuracy of the 2D pose estimator, so $w_{k} = c_k$. 
Third, as recent work~\cite{guo2017calibration} discovers that confidence scores from deep networks are not linear to the accuracy of the network, and  linearizing the scores is possible by employing  some calibration technique, transforming a confidence score $c_k$ to the  calibrated confidence score $c^*_{k}$. Thus, a third weight choice is the calibrated confidence score, where $w_{k} = c^*_{k}$. Although the second and third weight choices use confidence score to suppress the negative impact of wrong 2D pose estimation to a certain extent, without explicit control of the level of suppression, the 3D pose estimation is still inevitably affected by the 2D estimator's errors (i.e., 2D estimations with low confidence scores).

\begin{figure}
    \centering
    \includegraphics[width=7cm]{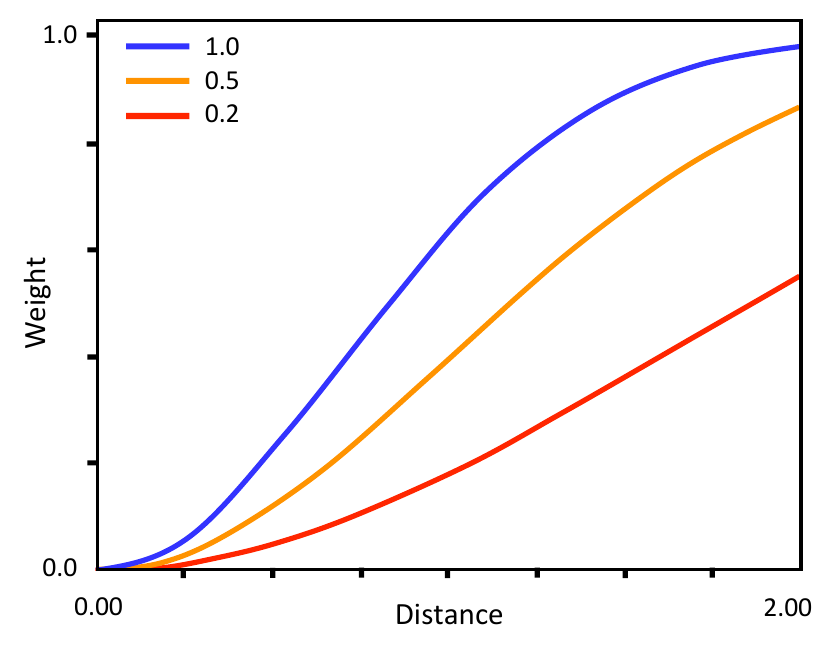}
    \caption{The plot of the weight function defined in Eq.~(\ref{eq:soft_weight}) using three different confidence values 1.0, 0.5, and 0.2, where $\sigma$ is set as $1$.}
    \label{fig:softthresh}
\end{figure}

We propose two approaches to suppress negative effect from false 2D estimations (those with low confidence scores). One is \textbf{hard-thresholding} that sets the $w_{k}$ as zero when confidence score is lower than a pre-defined threshold. The other approach is \textbf{soft-thresholding}, which is formulated as:
\begin{equation}
    w_{k} = 1 - exp(-\frac{c^*_k L_{2d}^2}{2 \sigma^2})
    \label{eq:soft_weight}
\end{equation}
where $\sigma$ is parameter to control the soft thresholding as a weight function. 
We visualize the weight curve with respect to the distance, $L_{2d}$, as in Figure~\ref{fig:softthresh} using different confidence values. 
At the same distance, $L_{2d}$, the higher the confidence (calibrated), the higher the weight, which indicates that the projection of estimated 3D pose would be enforced to match the output of 2D estimator stronger when the calibrated confidence is higher, leading to converge to smaller $L_{2d}$. On the other hand, if the 2D estimated keypoints have low confidence, its corresponding $w_{k}$ would be smaller, thus avoid the negative impact of wrong 2D estimation. Experiments in Table~\ref{tab:onlinelearning} illustrate the effectiveness of the proposed soft-thresholding strategy against hard-thresholding, or using constant, confidence or calibrated one directly as weight.

The overall inference stage optimization loss is a combination of re-projection loss $L_{rep}$, discrimination loss $L_{gen}$ as in Section \ref{sec:discriminator}, and a temporal smooth term to avoid turbulence in 2D estimation:
\begin{equation}
    L_{ISO} = L_{rep} + \lambda_{1} L_{gen} + \lambda_{2} \sum_t (X_{t+1} - X_t)^2
    \label{eq:iso_loss}
\end{equation}
where $\lambda_{1}$ and $\lambda_2$ are set to $0.1$ and $0.05$ in our experiments. 

\section{Experiment}

\subsection{Experiment Settings}

\begin{table}[b]
\footnotesize
\centering
  \begin{tabular}{c|c|c|c|c|c}
  \cline{1-6}
  \rule{0pt}{2.6ex}
  \textbf{Emb} & \textbf{T Len} & \textbf{T Strides}& \textbf{T Intvl} & P \#1 & P \#2 \\
    \cline{1-6}
    \rule{0pt}{2.6ex}
    64 & 64 & 1,2,3 & 1 & 58.3 & 44.2 \\
    128 & 64 & 1,2,3 & 1 & 46.7 & 36.1 \\
    256 & 64 & 1,2,3 & 1 & 43.1 & 33.8 \\
    512 & 64 & 1,2,3 & 1 & 42.6 & 33.4 \\
    1024& 64 & 1,2,3 & 1 & 42.9 & 33.6 \\
    \cline{1-6}
    512 & 8 & 1,2,3 & 1 & 50.2 & 40.1 \\
    512 & 16 & 1,2,3 & 1 & 46.9 & 36.0 \\
    512 & 32 & 1,2,3 & 1 & 44.0 & 33.9 \\
    512 & 64 & 1,2,3 & 1 & 42.6 & 33.4 \\
    512 & 128 & 1,2,3 & 1 & 42.9 & 33.7 \\
    \cline{1-6}
    512 & 64 & 1 & 1 & 45.4 & 35.9 \\
    512 & 64 & 1,2 & 1 & 44.3 & 34.8 \\
    512 & 64 & 1,2,3 & 1 & 42.6 & 33.4 \\
    512 & 64 & 1,2,3,5 & 1 & 41.8 & 32.1 \\
    512 & 64 & 1,2,3,5,7 & 1 & 41.2 & 31.5 \\
    \cline{1-6}
    512 & 64 & 1,2,3 & 1 & 42.6 & 33.4 \\
    512 & 64 & 1,2,3 & 3 & 43.1 & 33.7 \\
    512 & 64 & 1,2,3 & 5 & 44.0 & 34.4 \\
    \cline{1-6}
  \end{tabular}
  \caption{Parameter sensitivity test based on \textit{Protocol \#1 and \#2 of Human 3.6M dataset}. Emb stands for embedding dimension, T Len stands for Temporal length, T Strides stands for temporal strides, T Intvl stands for the temporal interval for TKCS.}
  \label{tab:sensitivity}
\end{table}

\textbf{Data Sets.} Human3.6M \cite{h36m_pami} is a large 3D human pose dataset. It has 3.6 million images including eleven actors performing daily-life activities, and seven actors are annotated. The 3D ground-truth is provided by the mocap system, and the intrinsic/extrinsic camera parameters are known. Similar to the existing methods  \cite{hossain2018exploiting,pavllo20183d,pavlakos2018ordinal,Yang20183DHP}, we use subjects 1, 5, 6, 7, 8 for training, and the subjects 9 and 11 for evaluation. 

HumanEva-I is a relatively smaller dataset. Following the typical protocol \cite{martinez2017simple,hossain2018exploiting,pavllo20183d}, we use the same data division to train one model for all three actions (Walk, Jog, Box), and use the remaining data for testing. 
MPI-INF-3DHP~\cite{mono-3dhp2017} is a relatively new dataset that is captured in an indoor setting which is similar to the setting of Human3.6M. Following recent methods ~\cite{kanazawa2018end,pavlakos2018ordinal,Chen_2019_CVPR} that report their performance on this dataset, we utilize this dataset for quantitative evaluation.
3DPW~\cite{3DPW} is a new dataset contains multi-person outdoor scenes. We use the testing set of 3DPW to perform quantitative evaluation following~\cite{martinez2017simple,humanMotionKanazawa19}. 

\textbf{Evaluation protocols.}
We apply a few common evaluation protocols in our experiments. \textit{Protocol \#1} refers to the Mean Per Joint Position Error (MPJPE) which is the millimeters between the ground-truth and the predicted keypoints. \textit{Protocol \#2}, often called P-MPJPE, refers to the same error after applying alignment between the predicted keypoints and the ground-truth. Percentage of Correct 3D Keypoints (3D PCK) under $150mm$ radius is used for quantitative evaluation for MPI-INF-3DHP following~\cite{mono-3dhp2017}. To compare with other human dynamics/pose forecasting methods, mean angle error (MAE) is used following~\cite{jain2016structural}. 

\subsection{Hyper-Parameter Sensitivity Analysis}
We conduct the sensitivity test of four hyper-parameters mentioned in this paper: embedding dimension for encoder, temporal length, temporal strides for TCN, and temporal interval for TKCS. The results are shown in Table~\ref{tab:sensitivity}. We find the best parameter settings by fixing three and adjusting the other one. To focus on understanding the influence of each parameter, semi-supervised learning using extra 2D data is disabled here. 

For the embedding dimension, we observe that within a reasonably large range, the performance is not affected significantly. The dimension $64$ is insufficient and results in large error. Within the range $256$ to $1024$, the errors only differ $0.4mm$, indicating that the model is insensitive to the setting of embedding dimension. 

For the temporal length, we test the range from $8$ to $128$. We can observe a steady reduction of errors until saturation at $128$. In addition, we adjust the temporal strides and find out that by adding more strides, the performance is improved and finally reaches $41.2mm$ with 5 strides compared to $45.4mm$ for single stride. We also test different temporal intervals for TKCS and observe interval $1$ produces the best performance. 

\begin{table}
\footnotesize
\centering
  \begin{tabular}{c|c|c}
  \cline{1-3}
  \rule{0pt}{2.3ex}
  \textbf{Method} & \textit{Protocol \#1} & \textit{Protocol \#2}  \\
    \cline{1-3}
    \rule{0pt}{2.6ex}
    Baseline (no ISO) & 41.2 & 31.5 \\
    Constant & 45.9 & 34.0\\
    Confidence & 43.0 & 32.7 \\
    Calibrated & 40.8 & 31.2 \\
    \cline{1-3}
    $th=0.3$ & 40.8 & 31.2  \\
    $th=0.5$ & 40.5 & 31.0  \\
    $th=0.7$ & 40.1 & 30.8  \\
    $th=0.9$ & 40.7 & 31.1  \\
    \cline{1-3}
    $\sigma = 1.0$ & 39.3 & 30.2 \\
    $\sigma = 0.5$ & 40.2 & 30.8 \\
    $\sigma = 0.3$ & 41.4 & 31.8 \\
    $\sigma = 2.0$ & 39.8 & 30.5 \\
    $\sigma = 3.0$ & 41.0 & 31.5 \\
    \cline{1-3}
  \end{tabular}
  \caption{Parameter analysis of different reprojection loss weight options on Human3.6M dataset. We compare the confidence score based weight choices (constant, confidence, and calibrated) with baseline (no ISO), as well as different parameter settings of hard-thresholding and soft-thresholding approaches.}
  \label{tab:onlinelearning}
\end{table}

Table~\ref{tab:onlinelearning} is an analysis of different options of weight function in ISO, 150 iterations of ISO are used for producing all the results.  In particular, the baseline means no ISO. Constant means no weight $w_k$ in Eq~\ref{eq:rep}, which can also be viewed as using a fixed constant weight. In this case, no matter $X_k^{det}$ is accurate or not, $L_{rep}$ always tries to minimize the difference between projected 3D keypoints and the estimated 2D keypoints. Confidence approach means directly using the confidence score as weight. Calibrated approach means linearizing the distribution of raw confidence score following~\cite{guo2017calibration}. Hard-thresholding results are listed from row 5 to 8 in the Table. Soft-threshold results are listed in the last five rows. Human3.6M dataset \textit{Protocol \#1} and \textit{\#2} are used for performance evaluation. Among the different weight function options, we observe that the confidence score based weight choices produce poor results in both \textit{Protocol \#1} and \textit{\#2}. Hard-thresholding results are better, but worse than the soft-thresholding results. Soft-thresholding result of $\sigma = 1.0$ is the best in the table and it is used in the rest of the paper for ISO experiments.

\begin{figure}[t]
\centering
\includegraphics[width=0.8\linewidth]{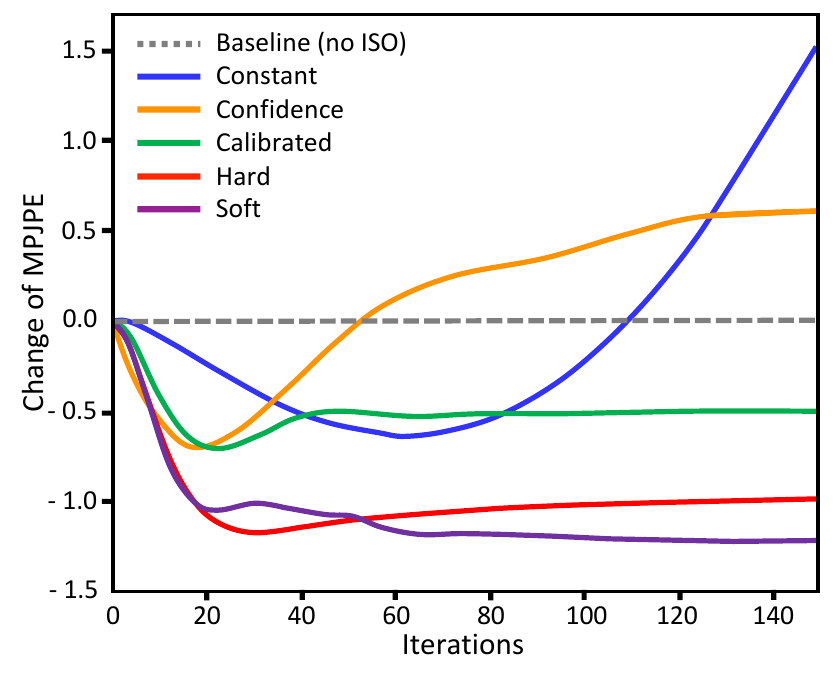}
\caption{MPJPE change of applying different re-projection weight options in ISO on one sample video. There are six curves plotted in total as explained in the legend. Baseline (no ISO) stands for no ISO is used, which is our offline trained model. Constant, confidence, calibrated stand for using constant, confidence score, and calibrated confidence score as weight. Hard and soft means hard-thresholding and soft-thresholding.} 
\label{fig:online_learning_curve}
\end{figure}

Figure~\ref{fig:online_learning_curve} shows the change of MPJPE curves where different weight functions for reprojection loss are used, we plot the MPJPE change over the 150 iterations of ISO update. For hard-thresholding weight option, the curve of $th=0.5$ is shown in Table~\ref{tab:onlinelearning} as it is the best hard-thresholding result, and we do the same for soft-thresholding where $\sigma = 1.0$. The curves in Figure~\ref{fig:online_learning_curve} provide a closer look to the behavior of using different weight options in the re-projection loss of ISO. As expected, if we perform ISO by ignoring the 2D estimator's error, it leads to the worst MPJPE increase as shown in the blue curve, because we enforce the projected 3D pose to match 2D estimation no matter the 2D result is accurate or not. Compared with other options, when the thresholding strategy is not applied, the negative impact of wrong 2D estimation is clearly observed after 30 iterations. Hard-thresholding is able to mitigate the problem while soft-thresholding can further reduce the negative effect on this video clip, which is consistent with the results in Table~\ref{tab:onlinelearning}.



\subsection{Ablation Study}

\begin{table}[h]
\footnotesize
\centering
  \begin{tabular}{c|c|c}
  \cline{1-3}
  \rule{0pt}{2.6ex}
  \textbf{Method} & \textit{Protocol \#1} & \textit{Protocol \#2} \\
    \cline{1-3}
    \rule{0pt}{2.6ex}
    Base & 51.7 & 40.5 \\
    +embedding & 51.0 & 40.1 \\
    +multi-stride TCN & 48.6 & 37.6 \\
    +Multi-view loss & 47.3 & 36.9 \\
    +Spatial KCS & 44.9 & 34.0 \\
    +Temporal KCS & 41.2 & 31.5\\
    +2D Data & 40.1 & 30.7\\
    +ISO & \textbf{38.8} & \textbf{29.7}\\
    \cline{1-3}
  \end{tabular}
  \caption{Ablation study on Human3.6M dataset under \textit{Protocol \#1 and \#2}. Best in bold.}
  \label{tab:ablation}
\end{table}

We conduct ablation studies to analyze each component of the proposed framework as shown in Table~\ref{tab:ablation}. 
As the baseline, we build a TCN to regress the 3D keypoints' positions based solely on their 2D coordinates $(x,y)$, which are obtained from the peaks in heatmaps from 2D pose detector. During TCN training, the 3D skeletons are also rotated along x, y, z axes as mentioned before. We use the standard MSE loss for the training.

We then add the modules one-by-one to perform ablation studies, including heatmaps embedding, multi-stride TCN, multi-view loss, spatial KCS, temporal KCS, 2D data semi-supervised learning, and ISO. We see that by adding more modules, the performance steadily improves, validating the effectiveness of our proposed modules. The largest improvements come from multi-stride TCN, spatial KCS, and temporal KCS modules. Temporal multi-scale features increase the capability of the networks to deal with videos with different speeds of motions. Although the spatial KCS  constraints the pose validity at individual frames properly, our temporal KCS clearly further improves the performance, which demonstrates that checking the pose validity of a single frame itself is insufficient, and checking the validity of the temporal pose sequence is necessary. Considering the fact that our offline trained model already achieves high accuracy, our ISO module can further improve the accuracy in both Protocol 1 and 2 by $1.3$ and $1.0 mm$, which demonstrates the effectiveness of the proposed self-supervised test time optimization.

\begin{table}[b]
\footnotesize
\centering
  \begin{tabular}{p{3.0cm}|p{0.15cm}p{0.18cm}p{0.4cm}|p{0.15cm}p{0.15cm}p{0.4cm}|p{0.4cm}}
  \cline{1-8}
  \rule{0pt}{2.6ex}
  \textbf{Method} & \multicolumn{3}{c|}{Walking} & \multicolumn{3}{c|}{Jogging} & Avg \\
    \cline{1-8}
    \rule{0pt}{2.6ex}Pavlakos et al. \cite{pavlakos2018ordinal}\MakeUppercase{*} & 18.8 & 12.7 & 29.2 & 23.5 & 15.4 & 14.5 & 18.3\\
    Hossain et al. \cite{hossain2018exploiting} & 19.1 & 13.6 & 43.9 & 23.2 & 16.9 & 15.5 & 22.0\\
    Wang et al. \cite{wang2019selfsupervised} & 17.2 & 13.4 & \underline{20.5} & 27.9 & 19.5 & 20.9 & 19.9\\
    Pavllo et al. \cite{pavllo20183d} & 13.4 & \underline{10.2} & 27.2 & \underline{17.1} & 13.1 & 13.8 & 15.8\\
    Cheng et al. \cite{ChengICCV19} & \underline{11.7} & \textbf{10.1} & 22.8 & 18.7 & \textbf{11.4} & \textbf{11.0} & \underline{14.3}\\
    \cline{1-8}
    Our result (offline) & \textbf{10.6} & 11.8 & \textbf{19.3} & \textbf{15.8} & \underline{11.5} & \underline{12.2} & \textbf{13.5}\\
    Our result + ISO & \textbf{10.8} & 11.5 & \textbf{19.0} & \textbf{15.1} & \underline{11.4} & \underline{11.6} & \textbf{13.2}\\
    \cline{1-8}
  \end{tabular}
  \caption{Evaluation on HumanEva-I dataset under \textit{Protocol \#2}. \textbf{Legend:} (*) uses extra depth annotations for ordinal supervision. Best in bold, second best underlined.}
  \label{tab:humanEvaI}
\end{table}

\begin{table}[h!]
\footnotesize
\centering
  \begin{tabular}{p{3.6cm}|p{0.4cm}}
  \cline{1-2}
  \rule{0pt}{2.6ex}
  \textbf{Method} & PCK \\
    \cline{1-2}
    \rule{0pt}{2.6ex}Mehta et al. 3DV \cite{mono-3dhp2017} & 72.5\\
    Yang et al. CVPR \cite{Yang20183DHP} &  69.0\\
    Chen et al. CVPR \cite{Chen_2019_CVPR} & 71.1\\
    Kocabas et al. CVPR \cite{Kocabas_2019_CVPR} & 77.5\\
    Wandt et al. CVPR \cite{Repnet} &  \underline{82.5}\\
    \cline{1-2}
    Our result (offline trained) & \textbf{84.1}\\
    Our result + ISO & \textbf{85.0}\\
    \cline{1-2}
  \end{tabular}
  \caption{Evaluation on MPI-INF-3DHP dataset using 3D PCK. Best in bold, second best underlined. Only overlapped keypoints with Human3.6M are used for evaluation.}
  \label{tab:3dhp}
\end{table}

\subsection{Quantitative Results}

The experiment results on Human 3.6M are shown in Table~\ref{tab:h3.6p1} and Table~\ref{tab:h3.6p2} for \textit{Protocol \#1} and \textit{\#2}, respectively. 
The result of our offline trained model reduces $2.8mm$ in MPJPE error compared to the state-of-the-art (SOTA)~\cite{ChengICCV19} and yields an $6.5\%$ error reduction. Our offline model also reduces $2.1mm$ in P-MPJPE and obtained $6.4\%$ error reduction against the SOTA~\cite{ChengICCV19}. The performance on actions which already have low error rates is not improved significantly, but for those actions such as photo capturing (abbreviated as Photo) and sitting down (abbreviated as SitD.), the errors are reduced by $>5mm$. Since in these actions, occlusion happens frequently, more temporal information and effective pose regularization are needed for producing correct estimations. Moreover, the result of our model using ISO reduces $4.2mm$ in MPJPE error and $3mm$ in P-MPJPE error against the SOTA and leads to $9.7\%$ and $9.1\%$ error reduction respectively. Considering existing methods almost get saturated on this dataset, our improvement is promising.

Table~\ref{tab:humanEvaI} shows our evaluation results on HumanEva-I dataset. $0.8mm$ and $1.1mm$ improvements are achieved on average by our offline trained model and ISO module, which implies an error reduction of $5.6\%$ and $7.7\%$ respectively. For MPI-INF-3DHP dataset, we  use  only our model trained on Human3.6M dataset, but do not fine-tune or retrain on the 3DHP data set. Following existing methods~\cite{chiu2019action,martinez2017human}, we evaluate the Percentage of Correct 3D Keypoints (3D PCK) where points error under $150mm$ is considered correct (as the keypoints definitions are different in Human3.6M and 3DHP, we evaluate only the overlapped keypoints). As shown in Table~\ref{tab:3dhp}, even if we do not perform any re-training or fine-tuning, we still achieve improvements of $1.9\%$ and $3\%$ in 3D PCK using our offline trained model and ISO module, indicating the effectiveness of our approach. 

\begin{table}[b]
\footnotesize
\centering
  \begin{tabular}{p{4cm}|p{1.2cm}}
  \cline{1-2}
  \rule{0pt}{2.6ex}
  \textbf{Method} & P-MPJPE \\
    \cline{1-2}
    \rule{0pt}{2.6ex}Mehta et al. 3DV'17 \cite{mono-3dhp2017} & 72.5\\
    Yang et al. CVPR'18 \cite{Yang20183DHP} &  69.0\\
    Chen et al. CVPR'19 \cite{Chen_2019_CVPR} & 71.1\\
    Kocabas et al. CVPR'19 \cite{Kocabas_2019_CVPR} & 77.5\\
    Wandt et al. CVPR'19 \cite{Repnet} &  \underline{82.5}\\
    Kolotouros et al. ICCV'19 \cite{kolotouros2019learning} & 59.2\\
    \cline{1-2}
    Our result (offline trained) & \textbf{66.4}\\
    Our result + ISO & \textbf{63.1}\\
    \cline{1-2}
  \end{tabular}
  \caption{Evaluation on 3DPW dataset using 3D P-MPJPE. Best in bold, second best underlined.}
  \label{tab:3dpw}
\end{table}

\begin{table*}[h]
\footnotesize
\centering
  \begin{tabular}{p{3.9cm}p{0.38cm}p{0.35cm}p{0.28cm}p{0.35cm}p{0.4cm}p{0.35cm}p{0.35cm}p{0.4cm}p{0.28cm}p{0.35cm}p{0.4cm}p{0.35cm}p{0.45cm}p{0.35cm}p{0.7cm}|p{0.4cm}}
  \cline{1-17}
    \rule{0pt}{2.6ex}
    \textbf{Method} & Direct & Disc. & Eat & Greet & Phone & Photo & Pose & Purch. & Sit & SitD. & Smoke & Wait & WalkD. & Walk & WalkT. & Avg\\
    \cline{1-17}
    \rule{0pt}{2.6ex}Fang et al. AAAI \cite{fang2018learning} & 50.1 & 54.3 & 57.0 & 57.1 & 66.6 & 73.3 & 53.4 & 55.7 & 72.8 & 88.6 & 60.3 & 57.7 & 62.7 & 47.5 & 50.6 & 60.4\\
    Yang et al. CVPR \cite{Yang20183DHP} & 51.5 & 58.9 & 50.4 & 57.0 & 62.1 & 65.4 & 49.8 & 52.7 & 69.2 & 85.2 & 57.4 & 58.4 & 43.6 & 60.1 & 47.7 & 58.6\\
    Hossain \& Little ECCV \cite{hossain2018exploiting} & 44.2 & 46.7 & 52.3 & 49.3 & 59.9 & 59.4 & 47.5 & 46.2 & 59.9 & 65.6 & 55.8 & 50.4 & 52.3 & 43.5 & 45.1 & 51.9\\
    Li et al. CVPR \cite{Li_2019_CVPR} & 43.8 & 48.6 & 49.1 & 49.8 & 57.6 & 61.5 & 45.9 & 48.3 & 62.0 & 73.4 & 54.8 & 50.6 & 56.0 & 43.4 & 45.5 & 52.7\\
    Chen et al. CVPR \cite{Chen_2019_CVPR} & - & - & - & - & - & - & - & - & - & - & - & - & - & - & - & 51.0 \\
    Wandt et al. CVPR \cite{Repnet} * & 50.0 & 53.5 & 44.7 & 51.6 & 49.0 & 58.7 & 48.8 & 51.3 & \underline{51.1} & 66.0 & 46.6 & 50.6 & \underline{42.5} & 38.8 & 60.4 & 50.9 \\
    Pavllo et al. CVPR \cite{pavllo20183d} & 45.2 & 46.7 & \underline{43.3} & 45.6 & 48.1 & 55.1 & 44.6 & 44.3 & 57.3 & 65.8 & 47.1 & \underline{44.0} & 49.0 & \textbf{32.8} & 33.9 & 46.8\\
    Cheng et al. ICCV \cite{ChengICCV19} & 
    \underline{38.3} & \underline{41.3} & 46.1 & \underline{40.1} & \underline{41.6} & \underline{51.9} & \underline{41.8} & \textbf{40.9} & 51.5 & \underline{58.4} & \underline{42.2} & 44.6 & \textbf{41.7} & 33.7 & \underline{30.1} & \underline{42.9}\\
    \cline{1-17}
    Our result (offline trained) & 
    \textbf{36.2} & \textbf{38.1} & \textbf{42.7} & \textbf{35.9} & \textbf{38.2} & \textbf{45.7} & \textbf{36.8} & \underline{42.0} & \textbf{45.9} & \textbf{51.3} & \textbf{41.8} & \textbf{41.5} & 43.8 & \underline{33.1} & \textbf{28.6} & \textbf{40.1} \\
    Our result + ISO & 
    \textbf{36.4} & \textbf{38.7} & \textbf{41.9} & \textbf{36.1} & \textbf{36.3} & \textbf{42.9} & \textbf{34.2} & \underline{40.4} & \textbf{43.2} & \textbf{49.7} & \textbf{40.7} & \textbf{40.0} & 43.2 & \underline{31.6} & \textbf{27.9} & \textbf{38.7} \\
    \cline{1-17}
  \end{tabular}
  \caption{Quantitative evaluation using MPJPE in millimeter between estimated pose and the ground-truth on Human3.6M under \textit{Protocol \#1}, no rigid alignment or transform applied in post-processing. Best in bold, second best underlined. * indicates ground-truth 2D labels are used.} 
  \label{tab:h3.6p1}
\end{table*}

\begin{table*}[h]
\footnotesize
\centering
  \begin{tabular}{p{3.9cm}p{0.38cm}p{0.35cm}p{0.28cm}p{0.35cm}p{0.4cm}p{0.35cm}p{0.35cm}p{0.4cm}p{0.28cm}p{0.35cm}p{0.4cm}p{0.35cm}p{0.45cm}p{0.35cm}p{0.7cm}|p{0.4cm}}
  \cline{1-17}
    \rule{0pt}{2.6ex}
    \textbf{Method} & Direct & Disc. & Eat & Greet & Phone & Photo & Pose & Purch. & Sit & SitD. & Smoke & Wait & WalkD. & Walk & WalkT. & Avg\\
    \cline{1-17}
    \rule{0pt}{2.6ex}Fang et al. AAAI \cite{fang2018learning} & 38.2 & 41.7 & 43.7 & 44.9 & 48.5 & 55.3 & 40.2 & 38.2 & 54.5 & 64.4 & 47.2 & 44.3 & 47.3 & 36.7 & 41.7 & 45.7\\
    Yang et al. CVPR \cite{Yang20183DHP} & \underline{26.9} & 30.9 & 36.3 & 39.9 & 43.9 & 47.4 & \textbf{28.8} & \underline{29.4} & \underline{36.9} & 58.4 & 41.5 & \textbf{30.5} & \textbf{29.5} & 42.5 & 32.2 & 37.7\\
    Hossain \& Little ECCV \cite{hossain2018exploiting} & 36.9 & 37.9 & 42.8 & 40.3 & 46.8 & 46.7 & 37.7 & 36.5 & 48.9 & 52.6 & 45.6 & 39.6 & 43.5 & 35.2 & 38.5 & 42.0\\
    Kocabas et al. CVPR \cite{Kocabas_2019_CVPR} & - & - & - & - & - & - & - & - & - & - & - & - & - & - & - & 45.0 \\
    Li et al. CVPR \cite{Li_2019_CVPR} & 35.5 & 39.8 & 41.3 & 42.3 & 46.0 & 48.9 & 36.9 & 37.3 & 51.0 & 60.6 & 44.9 & 40.2 & 44.1 & 33.1 & 36.9 & 42.6\\
    Wandt et al. CVPR \cite{Repnet} * & 33.6 & 38.8 & \underline{32.6} & 37.5 & 36.0 & 44.1 & 37.8 & 34.9 & 39.2 & 52.0 & 37.5 & 39.8 & 34.1 & 40.3 & 34.9 & 38.2 \\
    Pavllo et al. CVPR \cite{pavllo20183d} & 34.1 & 36.1 & 34.4 & 37.2 & 36.4 & 42.2 & 34.4 & 33.6 & 45.0 & 52.5 & 37.4 & 33.8 & 37.8 & \textbf{25.6} & \underline{27.3} & 36.5\\
    Cheng et al. ICCV \cite{ChengICCV19} & 28.7 & \underline{30.3} & 35.1 & \underline{31.6} & \underline{30.2} & \underline{36.8} & 31.5 & \textbf{29.3} & 41.3 & \underline{45.9} & \underline{33.1} & 34.0 & \underline{31.4} & 26.1 & 27.8 & \underline{32.8}\\
    \cline{1-17}
    Our result (offline trained) & \textbf{26.2} & \textbf{28.1} & \textbf{31.1} & \textbf{28.4} & \textbf{28.5} & \textbf{32.9} & \underline{29.7} & 31.0 & \textbf{34.6} & \textbf{40.2} & \textbf{32.4} & \underline{32.8} & 33.1 & \underline{26.0} & \textbf{26.1} & \textbf{30.7}\\
    Our result + ISO & \textbf{26.3} & \textbf{28.5} & \textbf{29.8} & \textbf{28.5} & \textbf{27.3} & \textbf{31.1} & \underline{28.3} & 29.9 & \textbf{33.3} & \textbf{38.5} & \textbf{31.3} & \underline{31.2} & 32.3 & \underline{24.7} & \textbf{25.4} & \textbf{29.8}\\
    \cline{1-17}
  \end{tabular}
  \caption{Quantitative evaluation using P-MPJPE in millimeter between estimated pose and the ground-truth on Human3.6M under \textit{Protocol \#2}. Procrustes alignment to the ground-truth is used in post-processing. Best in bold, second best underlined. * indicates ground-truth 2D labels are used.}
  \label{tab:h3.6p2}
\end{table*}

\begin{figure*}[h!]
\centering
\makebox[\textwidth]{\includegraphics[width=\textwidth]{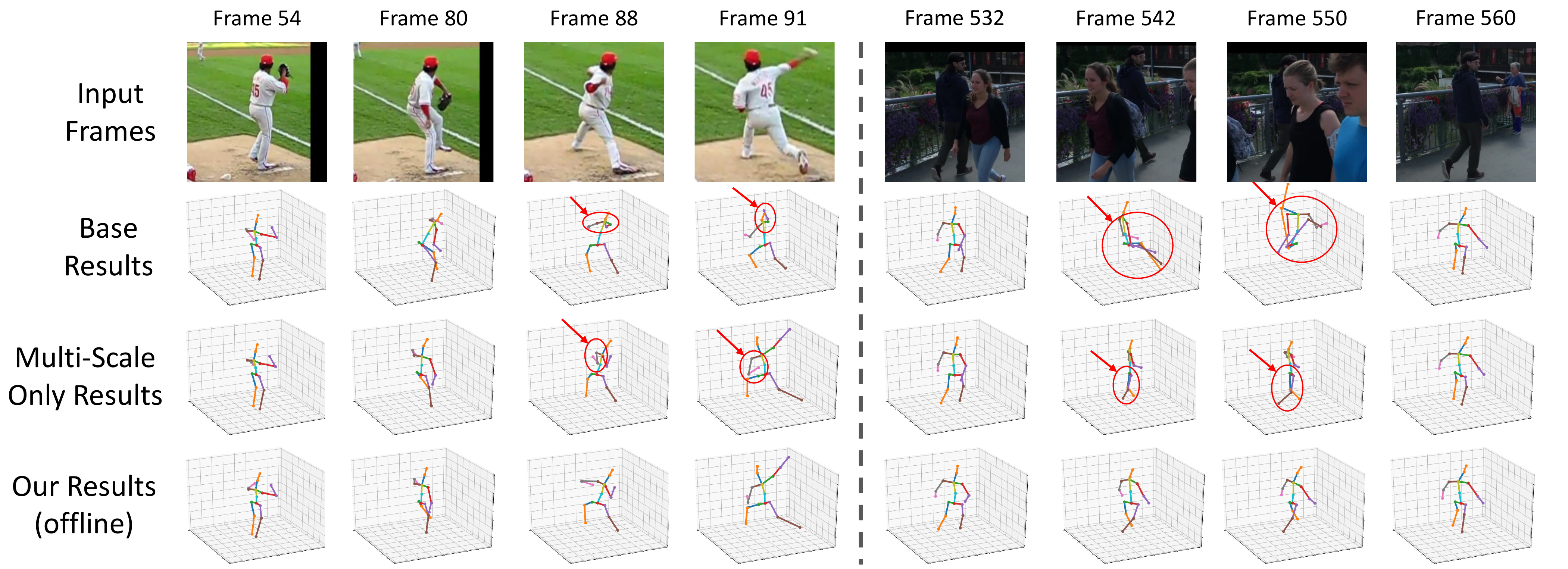}}
\caption{Examples of results from our offline framework compared with different baseline results. 
First row shows the images from two video clips;
second row shows the 3D results that uses baseline approach described in Ablation Studies; third row shows the 3D results that uses multi-scale temporal features without occlusion augmentation and spatio-temporal discriminator; last row shows the results of using all offline training modules including multi-scale temporal features, occlusion augmentation, and spatio-temporal discriminator. Wrong estimations are labeled in red circles.}
\label{fig:qualitative_evaluation}
\end{figure*}

As above 3D human pose datasets are mostly captured in single-person indoor scenes, we further evaluate our framework on 3DPW, a new outdoor multi-person 3D poses dataset. Following~\cite{martinez2017simple,humanMotionKanazawa19} we do not train on 3DPW and only use its testing set for quantitative evaluation. As the 3D ground-truth in 3DPW is defined using the SMPL format~\cite{SMPL2015}, which is different from the keypoint's definition of Human3.6M where our model is trained on, we perform a joint adaptation similar to~\cite{tripathi2020posenet3d} to transform our estimated 3D keypoints for a fair comparison. Quantitative comparison against other methods are summarized in Table~\ref{tab:3dpw}, which demonstrates our method is comparable to the SOTA. In particular, the result of~\cite{kolotouros2019learning} is better than ours because they used both Human3.6M and MPI-INF-3DHP as 3D ground-truth datasets in training their networks together with an additional model fitting process, but we only used Human3.6M in training our 3D network. 

\begin{table*}
\footnotesize
\centering
  \begin{tabular}{p{2.9cm}|p{0.25cm}p{0.25cm}p{0.25cm}p{0.25cm}p{0.5cm}|p{0.25cm}p{0.25cm}p{0.25cm}p{0.25cm}p{0.5cm}|p{0.25cm}p{0.25cm}p{0.25cm}p{0.25cm}p{0.5cm}|p{0.25cm}p{0.25cm}p{0.25cm}p{0.25cm}p{0.5cm}}
  \cline{1-21}
  \rule{0pt}{2.6ex}
  \textbf{Actions} & \multicolumn{5}{c|}{Walking} & \multicolumn{5}{c|}{Eating} & \multicolumn{5}{c|}{Smoking} & \multicolumn{5}{c}{Discussion}\\
  \cline{1-21}
  \rule{0pt}{2.6ex}
  \textbf{Milliseconds} & 80 & 160 & 320 & 560 & 1000 & 80 & 160 & 320 & 560 & 1000 & 80 & 160 & 320 & 560 & 1000 & 80 & 160 & 320 & 560 & 1000 \\
  \cline{1-21}
  \rule{0pt}{2.6ex}Ghosh et al. \cite{ghosh2017learning} & 1.00 & 1.11 & 1.39 & 1.55 & 1.39 & 1.31 & 1.49 & 1.86 & 1.76 & 2.01 & 0.92 & 1.03 & 1.15 & 1.38 & 1.77 & 1.11 & 1.20 & 1.38 & 1.53 & \textbf{1.73} \\
  Martinez et al. \cite{martinez2017human} & 0.32 & 0.54 & 0.72 & 0.86 & 0.96 & \underline{0.25} & 0.42 & 0.64 & 0.94 & 1.30 & 0.33 & 0.60 & 1.01 & 1.23 & 1.83 & 0.34 & 0.74 & 1.04 & 1.43 & 1.75 \\
  Chiu et al. \cite{chiu2019action} & \textbf{0.25} & \textbf{0.41} & \textbf{0.58} & \textbf{0.74} & \textbf{0.77} & \textbf{0.20} & \textbf{0.33} & \textbf{0.53} & \textbf{0.84} & \underline{1.14} & \textbf{0.26} & \underline{0.48} & \textbf{0.88} & \textbf{0.98} & \underline{1.66} & \textbf{0.30} & \underline{0.66} & \underline{0.98} & \underline{1.39} & \underline{1.74} \\
  Our result & \underline{0.29} & \underline{0.48} & \underline{0.65} & \underline{0.79} & \underline{0.92} & \underline{0.25} & \underline{0.39} & \underline{0.58} & \underline{0.87} & \textbf{1.02} & 0.34 & \textbf{0.44} & \underline{0.90} & \underline{1.07} & \textbf{1.52} & 0.33 & \textbf{0.63} & \textbf{0.90} & \textbf{1.30} & 1.77 \\
  \cline{1-21}
  \end{tabular}
  \caption{Evaluation on Human3.6M dataset on human dynamics protocol. Mean angle error of predicted 3D poses after different time intervals is used following~\cite{martinez2017human,ghosh2017learning}. The milliseconds is the set of future time for checking the performance. Best in bold, second best underlined.}
  \label{tab:dynamics}
\end{table*}

\begin{figure*}[h!]
\centering
\makebox[\textwidth]{\includegraphics[width=0.66\textwidth]{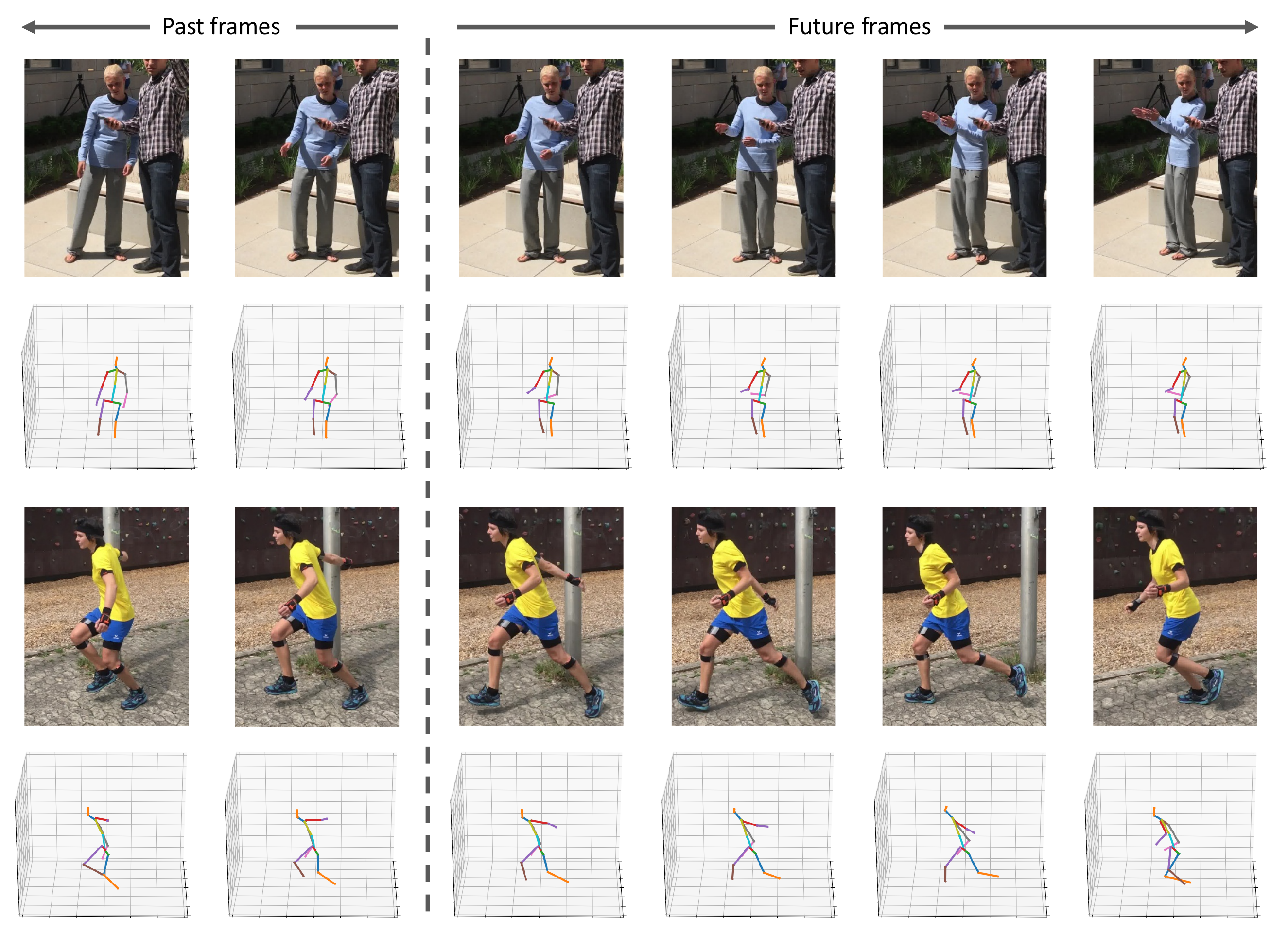}}
\caption{Examples of our pose prediction results. First and third rows are two video clips. Second and fourth rows are the corresponding results.}
\label{fig:pose_prediction}
\end{figure*}

We also evaluate our model's potential on human dynamics which is targeted to predict several future frames' 3D skeleton. The performance is shown in Table~\ref{tab:dynamics}. Note that, the SOTA method in human pose prediction actually uses past 3D ground-truth keypoints as input for prediction~\cite{chiu2019action}, while our method does not use any ground-truth but takes images from video as input to estimate the coordinate of the keypoints first, and then predict the future 3D information, which is a more difficult task. Nevertheless, we still achieve similar performance compared with the SOTA, which demonstrates the versatility of the proposed framework. Our method is not designed specifically for human pose prediction, but is a more generalized framework for pose estimation with or without observations (due to occlusions) in various scenarios. Figure~\ref{fig:pose_prediction} shows an example of our result on human pose prediction. Please note that in human pose prediction task, as there is no observation for future frames, we cannot apply the ISO module directly, so the results in Table~\ref{tab:dynamics} and Figure~\ref{fig:pose_prediction} are from our offline trained model without test time optimization.

\begin{figure}[h!]
    \centering
    \includegraphics[width=\linewidth]{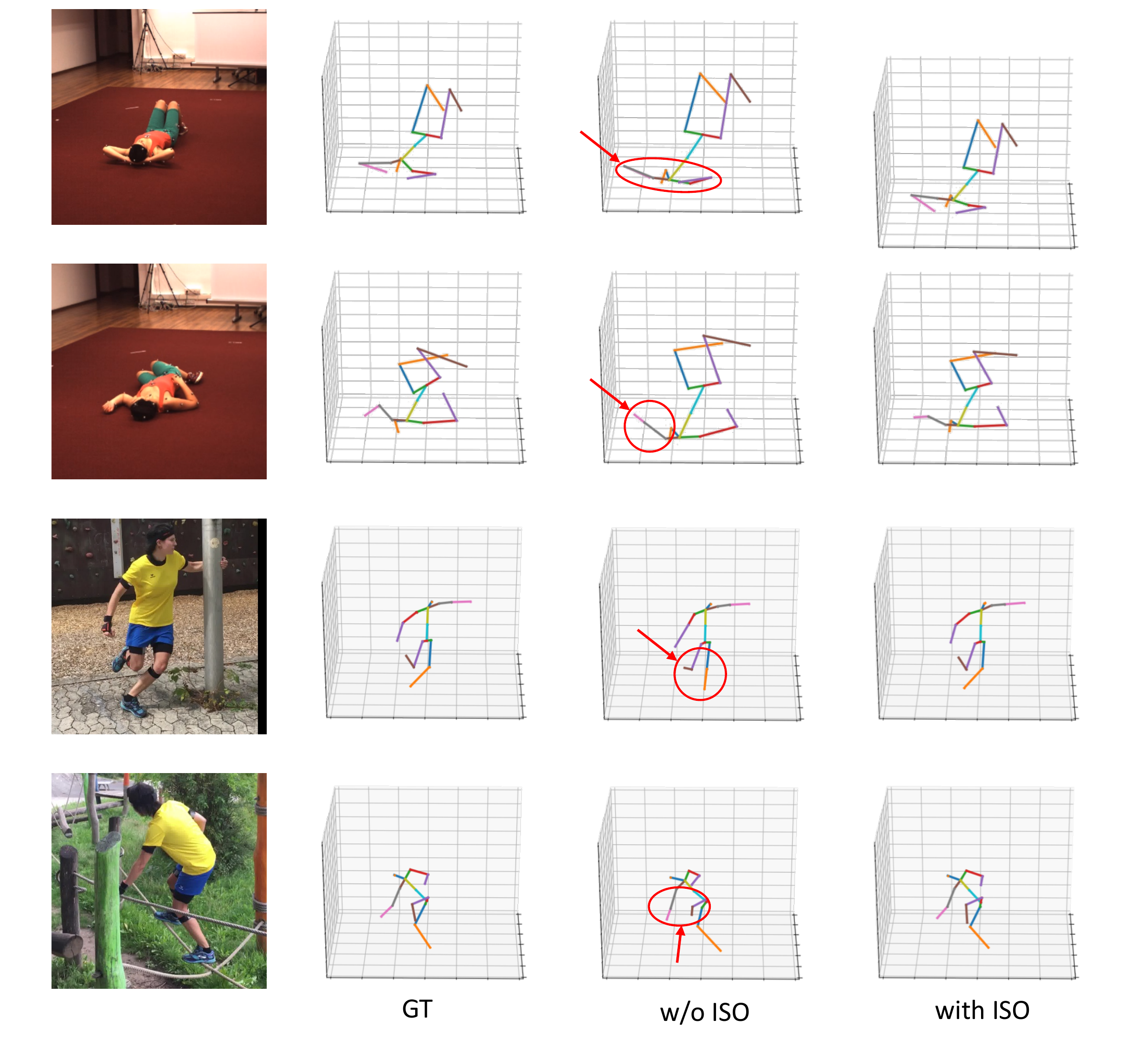}
    \caption{Qualitative results of ISO module. First two rows show results from H3.6M dataset, and last two rows show results from 3DPW dataset. Head-top keypoint does not exist in the ground-truth of 3DPW dataset, so it is omitted in the visualization of 3DPW results.}
    \label{fig:online}
\end{figure}

\subsection{Qualitative Results}

\begin{figure*}[t]
    \centering
    \includegraphics[width=0.8\textwidth]{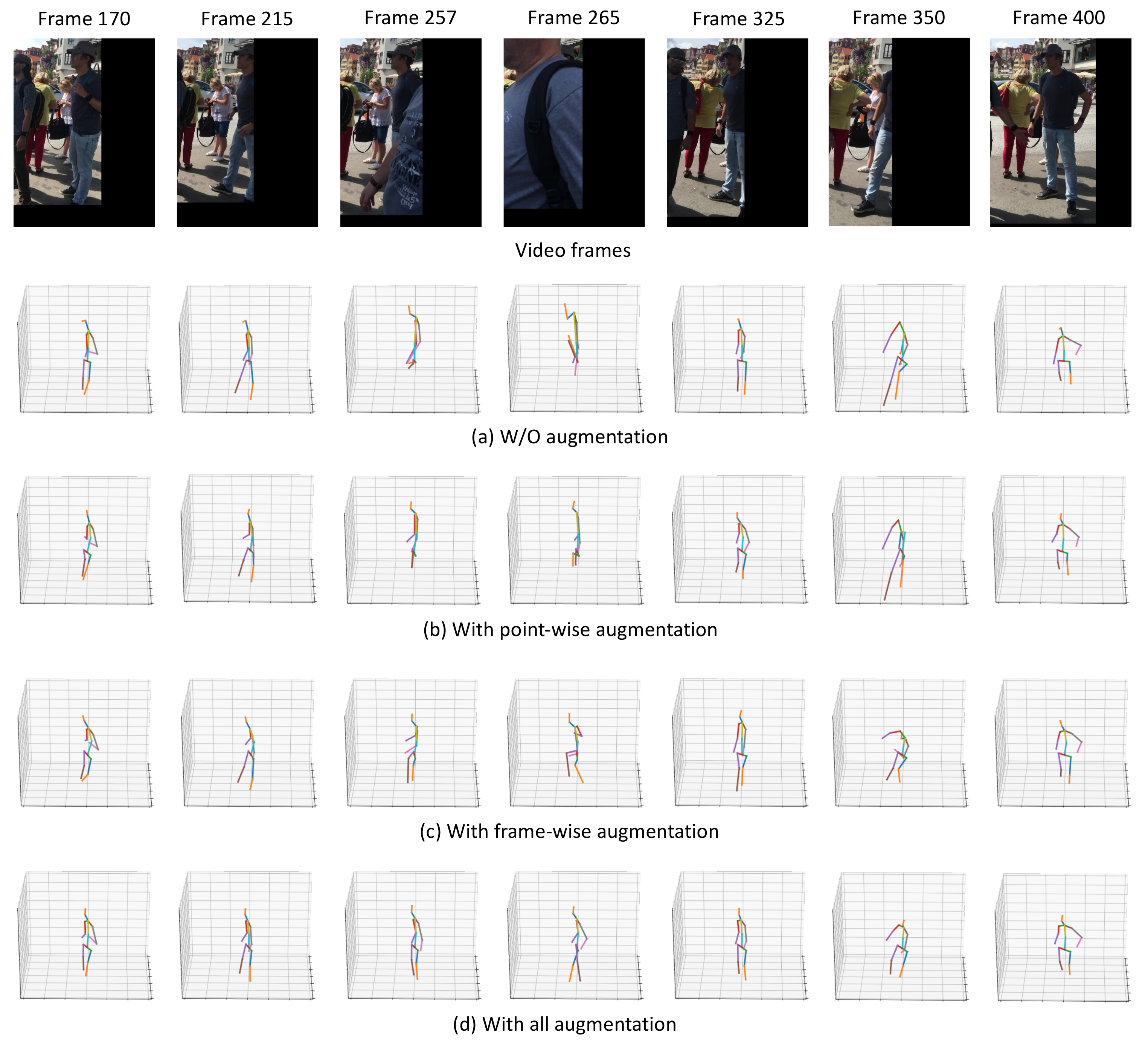}
    \caption{Qualitative results of using different occlusion augmentations. Row (a) are the 3D human pose results without any occlusion augmentation. The point-wise augmentation enables the model producing more accurate estimation with light occlusion as in frame 325 as in row (b). The frame-wise augmentation improves the estimation under severe occlusion as in frame 257 and 265 as in row (c). Due to the influence of missing 2D in previous frames (frame 215) and future frames (frame 325), merely using the frame-wise augmentation cannot solve the problem. After combining the two types of augmentation, the estimations are accurate and reasonable as in row (d).}
    \label{fig:augmentation}
\end{figure*}

Figure~\ref{fig:qualitative_evaluation} shows the 3D pose estimation results of the proposed framework compared with different baseline results. The first video clip (left four columns) shows a person playing baseball, which contains fast motion of limbs; the second video clip (right four columns) shows a person walking from left to right while some other people passing him, leading to occlusion. We use the same baseline method as used in the Ablation Study. The baseline (the second row) fails on both video clips, because it cannot handle fast motion or occlusion. The results in the third row are from the method that uses multi-scale temporal features but without occlusion augmentation and spatio-temporal KCS based discriminator. We observe that it can handle the fast motion case to some extent but fails on the occlusion video, and the generated poses do not always satisfy anthropometric constraints. The last row shows the results of our offline trained model and it demonstrates our method can handle different motion speeds and various types of occlusion. Figure~\ref{fig:online} provides insights regarding to when the proposed ISO module can help to improve the performance of the offline trained model. We can see the 3D human poses estimated by our offline trained model actually deviate the ground-truth, but the proposed ISO module is able to leverage the information from 2D estimator to perform self-supervised test time optimization to correct and improve the 3D pose estimation results. 

\subsection{Occlusion Augmentation}

We qualitatively evaluate models with different occlusion augmentation on videos captured in open environment. As shown in Figure \ref{fig:augmentation}, there are different kinds of occlusions may occur in videos: partial occlusion, fully occlusion and out-of-camera. According to the results, the model without occlusion augmentation is vulnerable under different types of occlusions. By adding the point-wise augmentation, the estimation is stabler for partial occlusion as in frame 325. With the frame-wise augmentation, the model is able to handle severe occlusion or even fully occlusion as in frame 257 and 265. The estimated pose may not be accurate but a reasonable pose is produced. The reason of inaccuracy is because the model is easily influenced by the missing or wrongly estimated points from previous or future frames (frame 215 and 325 in this case). Finally, using both kinds of augmentation is able to give reasonable predictions and follows human dynamics. 

\begin{figure*}[h!]
    \centering
    \includegraphics[width=\textwidth]{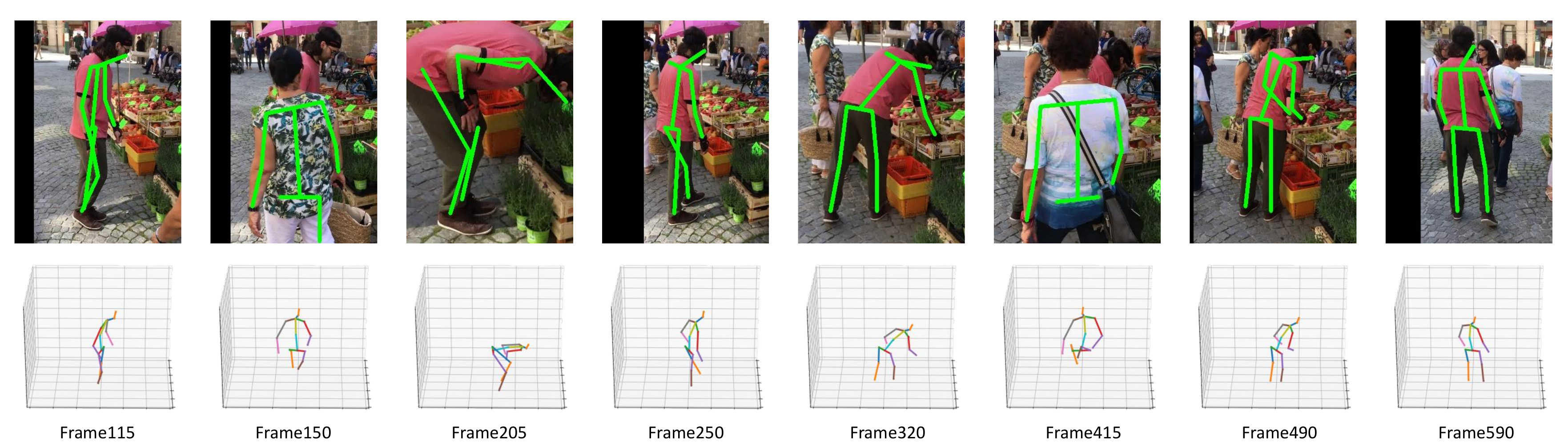}
    \caption{Some failure cases of our method. First row shows the estimation from 2D pose detector and second row shows the final 3D pose estimation results. Although our method is good at dealing with self-occlusion as in frame 115, 205, 250, 330 and 490, when our target is occluded by another person, the 2D pose detector will give undesired estimations which will mislead our 3D pose estimation, as shown in frame 150 and 415. }
    \label{fig:failurecase}
\end{figure*}

\section{Failure Cases}

Figure \ref{fig:failurecase} shows both 2D and 3D pose estimation results of our framework on a video sequence where failure takes place. Note that our model can deal with self-occlusion or occlusion by other non-human objects well in frame 115, 205, 250, 320, and 490. However, when there is an inter-person occlusion, the estimated poses are wrong. The reason is that our method is a top-down 3D human pose estimation framework, our 3D pose estimation depends on the 2D pose estimation, and the 2D pose estimation relies on object detection (i.e., detect object person bounding box). When two or more persons are close to each other, the detection bounding box of the target person may cover not only the target person but also others. Therefore, the estimated 2D keypoints may shift to another person, and then confuse the downstream 3D pose estimation process. Frame 150 and 415 are examples to show this type of failure as shown in Figure~\ref{fig:failurecase}, when the target person is occluded by another person (occluder), our 2D pose estimator gives predictions on the occluder instead, thus misleads our 3D pose estimation. Pose tracking or combining bottom-up and top-down frameworks are possible solutions to this type of failure, which are beyond the scope of this paper and we leave it as a future work.

\section{Discussion}


In this section, we discuss the need of applying our inference time optimization (ISO).
As our networks are trained with three types of datasets: 3D video datasets (with 3D pose ground-truth) like Human3.6M \cite{h36m_pami}, 2D video datasets (with 2D pose ground-truth) like Penn Action \cite{penndataset} for TCN and overall framework, and 2D image datasets (with 2D pose ground-truth) for 2D pose estimator like COCO \cite{lin2014microsoft} or MPII \cite{andriluka20142d}. The 3D datasets are usually collected in indoor or controlled environment due to it is costly to obtain the ground-truth, and the video datasets tends to have limited pose variations. In contrast, the 2D image datasets are easy to collect and relatively easy to annotate, therefore, the 2D image datasets have better coverage of pose variations as well as much more variety in appearance, background, and lighting conditions. As a result, there is a chance that the 2D pose estimator trained with the 2D image datasets can make correct pose prediction for unusual poses or challenging image appearance. Note that, there is no guarantee that the 2D pose estimator is always accurate, therefore, it cannot be simply used as pseudo ground-truth in test time optimization. 

The motivation of using Eq.(\ref{eq:rep}) is that we observe the 3D results from TCN does not match the keypoints of 2D estimation. Even we already have the Eq.(\ref{eq:l2d}) in training stage, the network's learning capability is still constrained by the limited video data. As the 2D image datasets like COCO or MPII cover a larger variation compared with the video datasets, the 2D pose estimator is potentially more robust and adaptive to pose variations under different scenarios. Therefore, we propose to optimize the 3D skeleton based on estimated 2D poses by taking the 2D pose estimator's confidence scores into account in the inference phase. 

The proposed ISO is naturally complementary to the offline training part of our framework. In training stage our multi-scale TCN and spatio-temporal pose discriminator take advantage of the information from video to achieve temporal smooth 3D pose estimation, but our method is also restricted by the lack of pose and appearance variations in video datasets. The proposed ISO allows us to take advantage of the rich pose and appearance variations in image datasets where the 2D estimator is trained on. In test time, the 2D pose estimator has the potential to produce reasonable estimation for unusual poses or challenging appearances, the ISO module enables us to optimize the 3D pose estimation based on the estimated 2D pose while taking the confidence scores of the 2D estimator into account. Therefore, our overall framework achieves the goal of taking advantage of both rich temporal information from 2D/3D video data and diverse pose and appearance information from 2D image data. As a result, our framework demonstrates superior performance on different benchmark datasets. 


\section{Conclusion}

In this paper, we present a new method based on four major components: multi-scale temporal features, spatio-temporal KCS pose discriminator, occlusion data augmentation, and inference stage optimization. Our method can deal with videos with various motion speeds and different types of occlusion. The effectiveness of each component of our method is illustrated in the ablation studies. To compare with the state-of-the-art 3D pose estimation methods, we evaluate the proposed method on four public 3D human pose datasets with commonly used protocols and demonstrate our method's superior performance. Comparison with the human dynamics methods is provided as well to show our method is versatile and potentially can be used for other human pose tasks, like pose forecasting.


\ifCLASSOPTIONcaptionsoff
  \newpage
\fi



%

\bibliographystyle{IEEEtran}
\bibliography{egbib}

%








\end{document}